\documentclass[a4paper,fleqn]{cas-sc}

\usepackage{float}
\usepackage[numbers]{natbib}
\usepackage{pifont}
\usepackage{threeparttable} 
\usepackage{tabularx}
\def\tsc#1{\csdef{#1}{\textsc{\lowercase{#1}}\xspace}}
\tsc{WGM}
\tsc{QE}
\begin{document}
\let\WriteBookmarks\relax
\def\floatpagepagefraction{1}
\def\textpagefraction{.001}

\shorttitle{}    

\shortauthors{Guangzhu Xu et~al.}  

\title [mode = title]{TransLPRNet: Lite Vision-Language Network for Single/double-line Chinese License Plate
Recognition}  

\tnotemark[1]



\author[1,2]{Guangzhu Xu}[type=editor,
                        auid=000,bioid=1,
                        style=chinese]

\affiliation[1]{organization={Hubei Key Laboratory of Intelligent Vision Based Monitoring for Hydroelectric Engineering, China Three Gorges University},
                addressline={}, 
                city={Yichang},
%               citysep={}, % Uncomment if no comma needed between city and postcode
                postcode={443002}, 
                state={Hubei},
                country={China}}

\author[2]{Zhi Ke}[style=chinese]

\author[2]{YunQi Wu}[style=chinese]

\author[2]{Pengcheng Zuo}[style=chinese]

\affiliation[2]{organization={College of Computer and Information Technology, China Three Gorges University}, 
                city={Yichang},
                postcode={443002}, 
                state={Hubei},
                country={China}}

\author[3,4]{Bangjun Lei}[style=chinese]

\affiliation[3]{organization={Hubei Key Laboratory of Digital Finance Innovation}, 
                city={Wuhan},
                postcode={430205}, 
                state={Hubei}, 
                country={China}}
\affiliation[4]{organization={School of Information Engineering, Hubei University of Economics}, 
                city={Wuhan},
                postcode={430205}, 
                state={Hubei}, 
                country={China}}


\begin{abstract}
License plate recognition in open environments has wide applications in many fields, but the diversity of license plate types and shooting conditions introduces numerous challenges and limitations. To address the problems existing in CNN and CRNN-based license plate recognition methods, this paper proposes a solution for single- and double-line Chinese license plate recognition that combines a lightweight visual encoder with a text decoder. To overcome the scarcity of double-line license plate image datasets, this paper constructs a dataset containing both single- and double-line plates by synthesizing images, pasting them onto real-world scenes, extracting them, and then mixing them with real license plate images. To further improve recognition accuracy, a perspective transformation-based license plate correction network (PTN) is proposed, which uses license plate viewpoint classification information as supervision and the regression values of license plate corner points as latent variables. This network offers better stability and interpretability with low annotation cost. On the corrected CCPD test set, the proposed algorithm achieves an average recognition accuracy of 98.75\% under coarse localization perturbations and 99.03\% under fine localization perturbations, demonstrating strong practical value.
\end{abstract}



% Keywords
\begin{keywords}
License Plate Recognition\\
Transformer\\
Visual-Language model\\
Perspective Transform Correction\\
Open Environment\\
\end{keywords}

\maketitle

% Main text
\section{Introduction}\label{}
Open license plate recognition (LPR) technology boasts a broad range of applications, including parking management, road traffic monitoring, toll station automation, and forensic evidence collection. Its primary advantage lies in the ability to operate without imposing additional constraints or restrictions on vehicles, making it applicable even when vehicles are in motion and captured at significant angles. However, in real-world open environments, LPR continues to face numerous challenges that demand resolution \cite{shi2023license}\cite{fan2022improving}\cite{liu2019vehicle}\cite{huang2020new}. These challenges stem from complex environmental lighting, variable weather conditions, and issues such as license plate soiling, particularly when the capture angle is uncontrolled.

Convolutional Neural Networks (CNNs) \cite{he2020robust} and Convolutional Recurrent Neural Networks (CRNNs) \cite{zou2022license} are widely adopted for extracting character features in license plate recognition. Via convolutional operations, these networks process characters in license plate images sequentially. As convolutional kernels are translated, a single character yields multiple feature outputs, each derived from image regions covered by different receptive fields. However, the tight arrangement of characters often leads to feature entanglement, resulting in the interweaving of character features within these regions. To address this issue, CRNN-based license plate recognition algorithms commonly incorporate Long Short-Term Memory (LSTM) or Bidirectional LSTM \cite{zou2020robust}networks. These architectures infer the most probable character class based on multiple local features extracted from each character, thereby improving recognition accuracy. Alternatively, pure CNN-based networks, such as LPRNet \cite{zherzdev1806lprnet}, employ global lateral convolution operations, enabling the model to capture more comprehensive character features and thus enhance understanding of inter-character relationships. Furthermore, these models often integrate the Connectionist Temporal Classification (CTC) loss function and dynamic decoding strategies \cite{hua2024recognition}, constructing a probability distribution space for character sequences. This approach resolves the automatic alignment of variable-length character sequences and further enhances recognition precision.

However, when license plate characters exhibit distortions due to camera angle, leading to variations in character size or glyph deformation, the aforementioned algorithms are prone to character omission or insertion errors, such as misidentifying a seven-character plate as an eight-character plate, or vice versa. This issue is particularly prevalent in license plate recognition schemes employing rectangular bounding boxes for localization \cite{raj2022license}. In contrast, license plate vertex localization techniques, which obtain the corner coordinates of the plate, enable more accurate license plate rectification \cite{adak2022automatic}. Nevertheless, this approach places a higher demand on training data annotation and the effectiveness of plate rectification is highly dependent on the precision of vertex localization.

To achieve adaptive license plate correction tailored for recognition tasks, an increasing number of license plate recognition schemes incorporate the integration of correction and recognition modules. Reference \cite{zherzdev1806lprnet} combines Spatial Transform Networks (STN) \cite{jaderberg2015spatial} with recognition networks to realize adaptive correction of license plate images through learned transformations. However, since STN relies on supervisory feedback from subsequent recognition networks to estimate affine transformation parameters, it requires pretraining of the recognition network, and only after reaching a certain performance level can it be integrated with the STN. Furthermore, due to the interdependent nature of perspective transformation parameters, employing STN for perspective correction often leads to issues such as divergence or non-convergence during training. Consequently, this approach proves difficult to adapt for license plate recognition in open environments with varying shooting angles.

On the other hand, current research on Chinese license plate recognition primarily focuses on single-line license plates, with limited attention given to double-line Chinese license plates \cite{deng2024collaborative}\cite{yang2024deep}. However, double-line plates are commonly used on vehicles such as trucks, buses, and trailers. This is partly due to the limitations of Convolutional Recurrent Neural Network (CRNN) architectures in processing vertical spatial sequence information, making it difficult to effectively model spatial dependencies in the vertical direction. Traditional Connectionist Temporal Classification (CTC) decoding relies on one-dimensional sequence features and cannot effectively encode the spatial relationships of characters in double-line plates. Furthermore, the lack of diverse license plate datasets, especially those containing double-line plate images, severely restricts the research and development of Chinese double-line license plate recognition.

With continual advancements in vision Transformer algorithms, the OCR solution based on the Transformer encoder-decoder architecture, TrOCR \cite{li2023trocr}, offers a novel approach to optical character recognition by leveraging its unique global self-attention mechanism. Compared to the limitations of traditional CNN/CRNN+CTC architectures in modeling one-dimensional sequences, the self-attention mechanism of Transformers can more effectively capture the relationships between arbitrary image regions within optical character images, thereby enhancing the accuracy of character recognition. Furthermore, Transformer text models pre-trained on large-scale unannotated datasets encode semantic priors, which can somewhat mitigate issues related to insufficient training data for double-line license plates. However, TrOCR suffers from disadvantages such as a large model size and low computational efficiency.

Building upon the aforementioned considerations and inspired by Spatial Transformer Networks (STN) and TrOCR, this paper introduces a novel license plate recognition framework that leverages a Transformer-based encoder-decoder architecture. We propose TransLPRNet, a lightweight model that integrates joint visual and textual pre-training for both single-line and double-line license plate recognition. Additionally, we design a License Plate Perspective Transformation Network (PTN), driven by view classification information that distinguishes whether an image depicts a frontal view of a license plate. An extended dataset is constructed by combining synthetically generated double-line license plate images with the CCPD dataset; this augmentation employs an information redundancy removal strategy to enrich the dataset with double-line plate images while preserving the completeness and scale of the original CCPD training set. Compared to other mainstream license plate recognition algorithms, our approach achieves superior results. The main contributions of this work are summarized as follows:

1. A lightweight vision-language hybrid license plate recognition network, TransLPRNet, is proposed. It employs a pre-trained MobileViTv3 \cite{wadekar2022mobilevitv3} model as the encoder and a lightweight transformer decoder. Experimental results show that, compared to TrOCR, TransLPRNet significantly reduces the number of model parameters and computational complexity, while also achieving notable improvements in recognition accuracy and inference speed. Additionally, this network demonstrates excellent performance and strong adaptability in recognizing single-line and double-line Chinese license plates.

2. A Perspective Transformation Network (PTN) for automatic rectification is introduced, which leverages weak supervision from a lightweight binary classification network to enable rapid automatic correction of various license plate types. PTN effectively rectifies perspective distortions in license plates captured from different scenes by identifying frontal-view images, thereby significantly reducing annotation costs.

3. To address the lack of double-line license plate datasets in unconstrained environments, this study employs a texture-mapping approach, overlaying generated double-line license plates onto real scenes and mixing them with authentic license plate images. This method constructs a diverse double/single-line license plate dataset. Furthermore, by compressing redundant information in the original CCPD dataset without increasing the number of training samples and while preserving all original data, the study extends the CCPD dataset to include images of double-line license plates.

The remainder of this paper is organized as follows. Section 2 summarizes the advantages and disadvantages of existing license plate recognition and correction algorithms, and introduces relevant publicly available datasets. Section 3 presents the proposed algorithms and the dataset constructed in this study. Section 4 reports the experimental results and provides a detailed analysis. Finally, the conclusion discusses the key findings and outlines future research directions.
\section{Related work}\label{}
\subsection{License Plate Recognition}\label{}
For license plate recognition (LPR) tasks in open environments, current mainstream algorithms often employ OCR networks based on CNNs and CRNNs (CNN+RNN). Zherzdev et al. \cite{zherzdev1806lprnet} pioneered LPRNet, a real-time LPR model that dispenses with the RNN component. This model utilizes a lightweight CNN backbone and replaces traditional LSTMs with wide convolutions for capturing local contextual information of characters. Coupled with the Connectionist Temporal Classification (CTC) loss function \cite{hua2024recognition}, LPRNet directly outputs variable-length character sequences. Their method achieved an average recognition accuracy of   95\% on a Chinese LPR dataset, demonstrating a favorable balance between recognition speed and accuracy. Xu et al. \cite{xu2018towards}constructed and open-sourced CCPD, the first comprehensive LPR dataset covering complex scenarios. They also proposed RPNet, a unified network architecture that integrates license plate detection and character recognition into an end-to-end pipeline. In the recognition stage, RPNet employs CRNN (CNN + BiLSTM + CTC) to directly recognize character sequences from Regions of Interest (ROIs) regressed from the detection box.

However, these methods often encounter significant challenges when processing double-line license plates. Traditional Convolutional Neural Network (CNN) or Convolutional Recurrent Neural Network (CRNN) architectures typically struggle to effectively handle multi-line information, particularly when dealing with uneven character distributions and variable numbers of lines. This limitation impedes the network's ability to capture contextual dependencies, consequently hindering the accuracy and robustness of double-line license plate recognition in complex scenarios. To address this limitation, Qin et al. \cite{qin2020efficient} proposed a unified framework for recognizing both single-line and double-line license plates. Their approach leverages an improved lightweight CNN for efficient feature extraction and employs a multi-task learning strategy to simultaneously perform license plate classification and character recognition. Finally, the recognition task is formulated as a sequence labeling problem and solved using Connectionist Temporal Classification (CTC) loss. While effective for some cases, these methods often require segmenting the double-line license plate into upper and lower regions and subsequently integrating the individual recognition results. Consequently, their performance degrades significantly when processing highly skewed or tilted license plates, often leading to the omission of some characters during recognition.

Compared to traditional OCR methods, deep learning-based OCR models like PaddleOCR \cite{du2020pp} have significantly improved recognition accuracy and computational efficiency through optimized algorithms and the incorporation of novel techniques, particularly excelling in the processing of double-line license plates. Li et al. \cite{li2022pp}proposed PP-OCRv3, which, building upon its predecessor, introduces several enhancements, including the lightweight SVTR-LCNet recognition network, an attention-guided CTC training strategy, and diverse data augmentation methods, enabling a dynamic balance between accuracy and speed.

Although the aforementioned lightweight solutions effectively mitigate the deployment efficiency bottlenecks of traditional models, their backbone architectures still rely on CNNs, which limits their capacity to model long-range dependencies and hampers the comprehensive representation of complex semantic and contextual relationships between characters. Additionally, such approaches face challenges in recognizing double-line license plates. To further enhance the expressive power and robustness of recognition models, researchers have recently begun exploring unified vision-text modeling frameworks based on Transformer architectures. These methods employ collaborative modeling between visual encoders and text decoders, incorporating large-scale pre-training and fine-tuning mechanisms to achieve stronger sequence modeling and cross-modal feature alignment. For example, the end-to-end text recognition model TrOCR proposed in \cite{li2023trocr} utilizes collaborative image and text Transformers to generate complex text sequences, thereby improving the modeling capability for intricate character sequences. However, Transformer-based recognition methods generally depend on large pre-trained models, which demand substantial computational resources and have slower inference speeds, presenting significant challenges for deployment in real-time recognition applications.

In recent years, with the development of lightweight vision-language Transformer models, networks such as MobileViTv3 \cite{mehta2021mobilevit} and lightweight transformer decoder  have demonstrated promising performance and deployment efficiency across various vision and text modeling tasks. Specifically, MobileViT achieves efficient encoding of image structural information by integrating the local perception of CNNs with the global modeling capability of Transformers. Building on this, we propose a lightweight end-to-end license plate recognition network architecture that jointly utilizes MobileViTv3 \cite{wadekar2022mobilevitv3} and lightweight transformer decoder, aiming to simultaneously balance recognition accuracy and deployment efficiency, making it suitable for real-time license plate recognition tasks in resource-constrained scenarios.
\subsection{ License Plate Spatial Transformation Correction}\label{}
Current OCR architectures have achieved significant progress in license plate recognition tasks; however, license plate images captured in real-world, open environments often exhibit issues such as angular tilt and geometric distortions, which pose challenges to subsequent character recognition in terms of accuracy and robustness. To improve recognition performance, license plate image rectification has increasingly become a critical preprocessing step and an active area of research.

Currently, the mainstream approaches for license plate rectification can be broadly categorized into two types. The first involves obtaining the four corner coordinates of the license plate region through a plate localization module, and then computing a perspective transformation matrix based on these coordinates and the desired rectified image size to achieve rectification. For example, Kundrotas and colleagues \cite{khokhar2024integrating} proposed a lightweight network to detect the four corners of the license plate and employed a perspective inverse transformation to perform geometric correction, thereby simplifying subsequent character recognition tasks. Their method utilizes an improved Hourglass network as a feature extractor and achieved an average recognition accuracy of 96.19\% on a Chinese license plate dataset. However, such approaches rely heavily on high-quality corner annotations and treat the rectification and recognition modules as separate entities. This separation makes it challenging to optimize the entire process end-to-end for stable spatial correction aimed at license plate character recognition.

The second category encompasses methods leveraging learnable Spatial Transformer Networks (STNs) \cite{jaderberg2015spatial}. These approaches employ a machine learning paradigm, adjusting network parameters by backpropagating the error signal derived from the subsequent license plate recognition network. In contrast to the first category, these methods obviate the need for manual annotation of corner points, thereby exhibiting enhanced adaptability. Xiao et al. \cite{xiao2021robust} utilized the YOLOv2 detector for license plate detection and proposed the ICSTN-CRNN model. This model integrates a Thin-Plate Spline-based Spatial Transformer Network (STN) to achieve automatic license plate rectification and recognition, demonstrating robust performance across multiple datasets. Furthermore, Akshay Bakshi et al. \cite{bakshi2023alpr} adopted a hybrid approach, combining Spatial Transformer Networks (STNs) with Convolutional Neural Networks (CNNs). They proposed an automatic license plate recognition system capable of rectification and character recognition under complex environmental conditions and multi-angle captures, achieving high recognition accuracy across datasets encompassing diverse regions and varying acquisition conditions.

Although the aforementioned methods partially alleviate issues related to angular deviation and deformation, the Spatial Transformer Networks (STNs) employed in these algorithms are primarily limited to spatial correction involving affine transformations. However, in open-world environments, license plate images often exhibit perspective distortions, rendering affine-based STNs insufficient for effective rectification. While the solution proposed in reference \cite{xiao2021robust} can address perspective transformation challenges by regressing 110 points, the large number of regression points may lead to model overfitting. This, in turn, can cause the model to exhibit instability and prediction fluctuations during the testing phase.
\subsection{ License Plate Datasets}\label{}
The license plate dataset is a crucial component of license plate recognition tasks. A high-quality, diverse Chinese license plate dataset provides a solid foundation for model training and evaluation. Currently, the main Chinese license plate datasets include CCPD \cite{xu2018towards}, CLPD \cite{zhang2020robust} and LSV-LP \cite{wang2022lsv}. Among these, the CCPD dataset is the most widely used Chinese license plate recognition dataset, comprising approximately 300,000 labeled single-line license plate images, which are divided into multiple subsets based on environmental variations such as Rotation, Tilt, Blur, and Weather conditions. The CLPD dataset, released by the Institute of Automation, Chinese Academy of Sciences, contains 1,200 license plate images representing various provinces across China. Due to its relatively small size and limited scenarios, its application scope is more constrained. LSV-LP is a Chinese license plate recognition dataset containing approximately 400,000 images, covering various license plate types and common noise factors, and is frequently used to evaluate the performance of recognition algorithms in complex scenes. Compared to CCPD, LSV-LP still exhibits certain deficiencies in annotation accuracy, image clarity, and coverage of specialized scenarios.

Although current Chinese license plate datasets offer a foundation for training and evaluation of recognition algorithms, certain shortcomings necessitate further attention. A significant limitation stems from label errors within some datasets. Because these datasets are annotated manually, a degree of mislabeling is unavoidable. Another critical deficiency is the insufficient proportion of double-line license plates. Given the widespread use of double-line plates in practice, this imbalance makes it challenging to effectively evaluate the performance of license plate recognition algorithms specifically on this important class of license plates.
\section{Our method}\label{}
\begin{figure}[h]%[]
  \centering
    \includegraphics[width=1\linewidth]{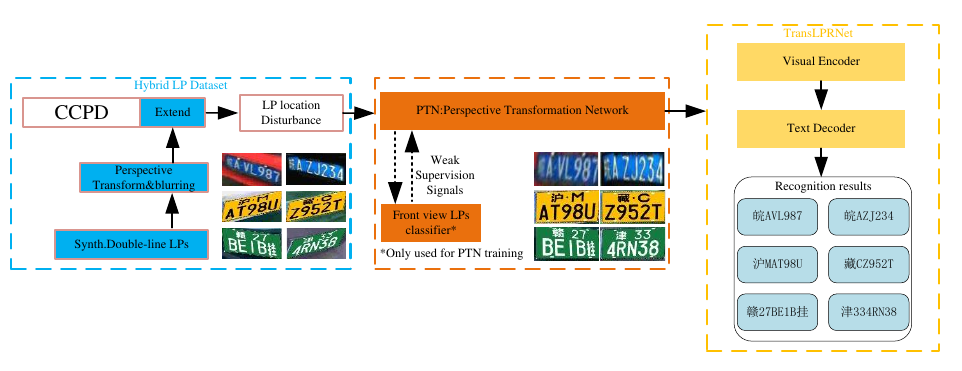}
    \caption{System solution diagram}\label{fig1}
\end{figure}

To address the challenges of license plate recognition in open environments, this paper first introduces a novel dual-style license plate recognition network, TransLPRNet. The model leverages a lightweight visual encoder, MobileViTv3 \cite{wadekar2022mobilevitv3}, combined with a lightweight transformer decoder, to enhance feature extraction and recognition accuracy. To further improve performance, a versatile perspective transformation space auto-correction network, PTN, is designed to correct images captured from various angles and deformations. Additionally, considering the lack of non-constrained environment datasets for dual-style license plates, we propose a dataset construction method. This involves synthesizing dual-style plates using templates and blur transformations, then seamlessly replacing the original license plates in redundant images within the CCPD dataset. These redundant images are highly similar and contain limited unique information, ensuring minimal impact on the overall dataset diversity. This approach enables the construction of suitable datasets for non-constrained environment recognition. Refer to Figure \ref{fig1} for the system architecture.
\subsection{ TransLPRNet}\label{}
\begin{figure}%[]
  \centering
    \includegraphics[width=0.6\linewidth]{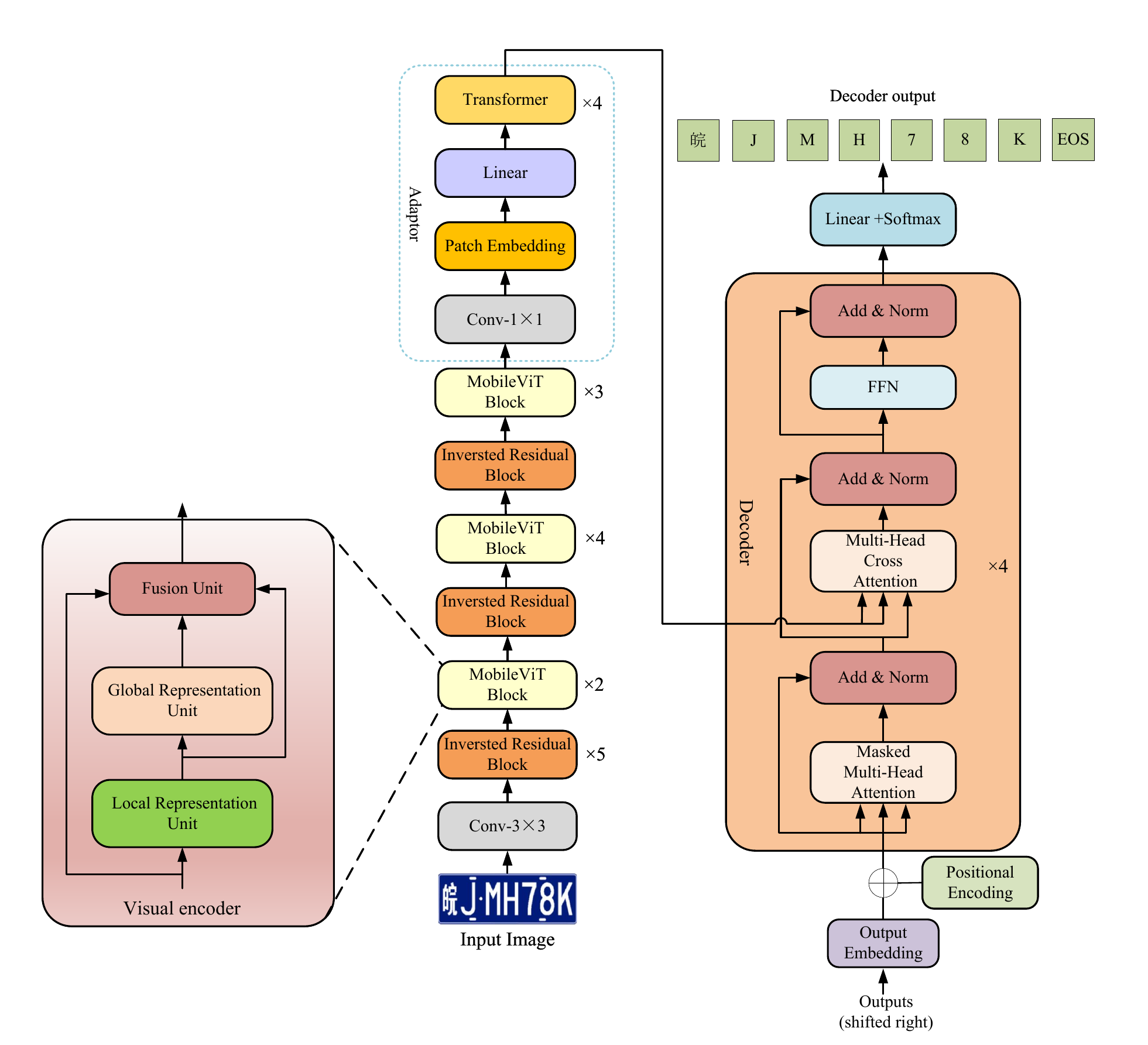}
    \caption{TransLPRNet network structure diagram }\label{fig2}
\end{figure}

The network structure of TransLPRNet is shown in Figure \ref{fig2}, which mainly consists of three key components: a visual encoder, an adapter, and a text decoder. The visual encoder employs a hybrid structure that alternately connects multiple IRB and MobileViT modules, forming a complementary feature extraction mechanism. IRB, with its lightweight characteristics, efficiently captures local details of license plate characters, providing foundational features for accurate single-character recognition. Meanwhile, the MobileViT module, through its window-based self-attention mechanism, achieves efficient modeling of the global layout of license plate characters while effectively controlling computational complexity. The adapter is responsible for matching the number of tokens and dimensions between the output of the visual encoder and the input of the text decoder. This decoupled "encoder-adapter-decoder" architecture offers the advantages of enhanced global semantic understanding and structural configurability.
\subsubsection{ TransLPRNet image encoder and adapter}\label{}
\begin{figure}%[]
  \centering
    \includegraphics[width=0.8\linewidth]{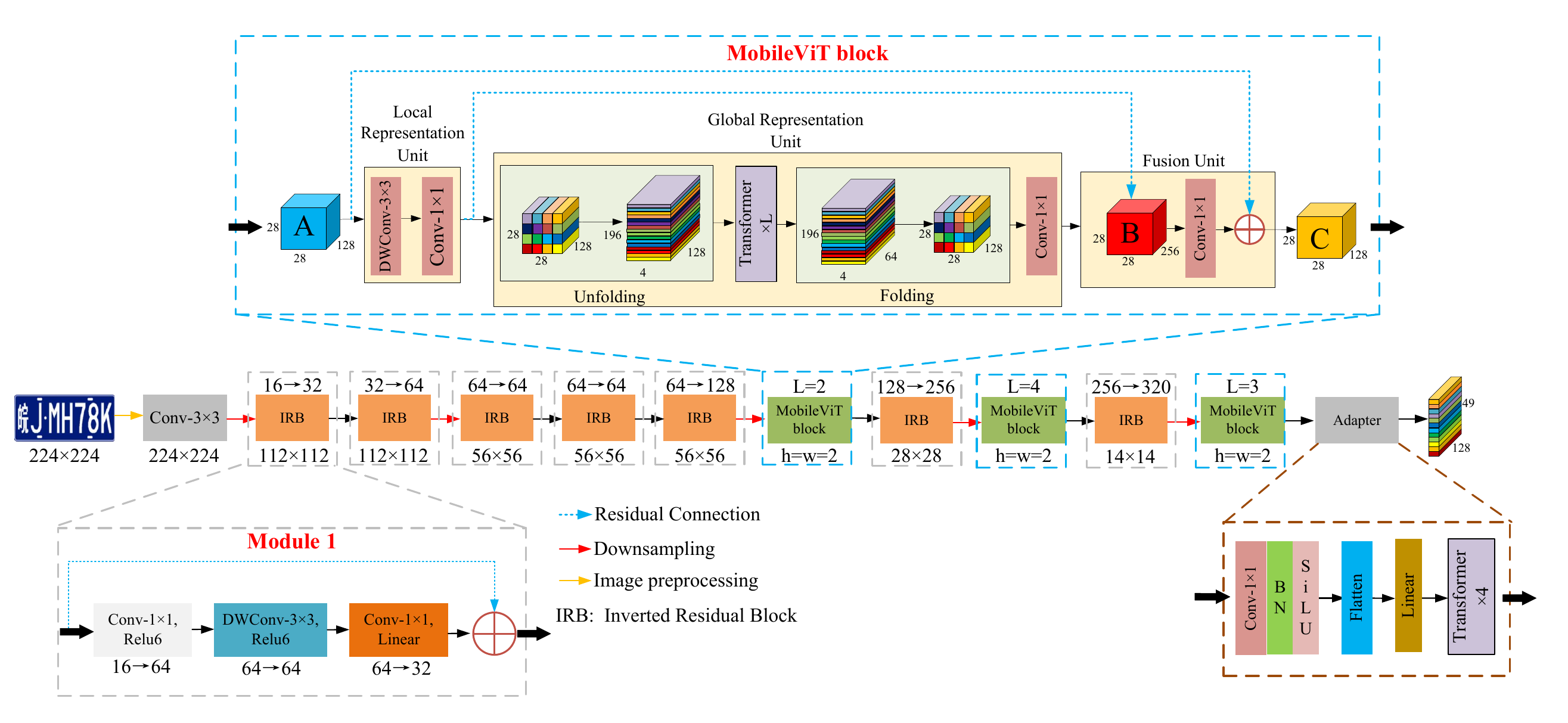}
    \caption{TransLPRNet encoder network structure diagram}\label{fig3}
\end{figure}
Figure \ref{fig3} illustrates the composition structure of the license plate visual encoder and adapter modules in the TransLPRNet model adopted in this paper. Among them, IRB is an inverted residual block derived from MobileNet\cite{howard2019searching}. This module first expands the channel dimension and then compresses it, while utilizing a depthwise separable convolution structure, ultimately achieving efficient local feature extraction. It is widely used in MobileViT\cite{wadekar2022mobilevitv3}.

Taking the first IRB, marked by the gray dashed outline, as an example: it begins with a 1×1 convolution that expands the number of channels from 16 to a higher dimension (specifically, 16 multiplied by the expansion ratio), which enhances the feature representation. In MobileViTv3, the expansion ratio for all IRB is 4. Next, a 3×3 depthwise convolution layer extracts spatial information. Finally, a 1×1 convolution reduces the number of channels to 32, which is the output. Since the first IRB has different input and output channel numbers, MobileViTv3 omits the residual connection (shown by the blue dashed arrow inside the gray dashed box) to reduce computation associated with downsampling.

In this encoder, the first convolutional layer within IRB 2, 5, 6, and 7 has a stride of 2, resulting in downsampling. The 3rd and 4th IRB are identical cascade modules; their internal first convolution layers have a stride of 1, and since their input and output channels are the same, residual connections are used in these blocks (as shown by the blue dashed arrow in module 1 in the figure).

As indicated by the blue dashed box in Figure 3, the MobileViTv3 module is composed of three main parts: a local representation unit, a global representation unit, and a fusion module. The local representation unit is designed to capture pixel-level local features while retaining fine spatial details, which it achieves using depthwise separable convolutions. To mitigate the computational cost of the ensuing Transformer, it further employs 1×1 convolutions for channel dimension compression.

The Global Representation Unit (GRU) is designed for efficient processing of feature information within local windows. Taking the first MobileViTv3 block in Figure 3 as an example, it receives a feature map of size 28×28×128. This feature map is first divided into multiple 2×2 local windows and then unfolded. Each 2×2 local window generates four tokens, each with a dimension of 128. The entire feature map contains 784 such local windows (28 × 28 / (2 × 2) =196), resulting in a total of 784 tokens (196 × 4 = 784). These tokens are grouped into 196 sets, with each set comprising four tokens corresponding to an original local window.

The core of the GRU lies in its four parallel Transformer layers. These layers iteratively perform Self-Attention computations on the four 128-dimensional tokens within each independent local window. This design enables the model to achieve pixel-level global correlation modeling within a small scope (i.e., inside each local window), effectively capturing complex dependencies within local regions.

Upon completion of computations across all local windows, the Transformer-processed tokens are folded back to the original spatial dimensions of the feature map (28×28×128). Finally, a 1×1 convolutional layer is applied to increase the channel dimension of the features, thereby enhancing the model's feature representation capability and providing richer semantic information for subsequent modules.

The Fusion Unit initially concatenates the outputs of the Local and Global Representation Units and employs a 1×1 convolution to achieve feature fusion. Subsequently, the fused features are added to the input of the MobileViTv3 module, forming a residual connection. Across different MobileViTv3 modules, interaction between local windows is facilitated via a 3×3 convolution within the inverted residual structure, effectively enabling the modeling of global information.

To meet the dual demands of accuracy and speed in license plate recognition, we modified the original MobileViTv3 architecture. Specifically, we removed its global average pooling layer and fully connected layers, which are typically used for image classification, and utilized the remaining components as the backbone for our visual encoder's feature extraction. To accelerate training, we leveraged the pre-trained weights of this backbone on the ImageNet-1k dataset.

The output of the last MobileViT in Figure \ref{fig3} is the output of the encoder, with a dimension of 7×7×320. Considering the token dimension requirements of the subsequent decoder, this paper introduces an adapter to bridge the visual encoder and the text decoder. In the adapter module, a 1×1 convolution is first used to reduce the channel dimension from 320 to 128, resulting in a 7×7×128 feature map. This feature map is then flattened into a 49×128 token sequence and normalized through a linear layer. Finally, these tokens are fed into a 4-layer Transformer encoder (each layer contains 8 attention heads and a feedforward network dimension of 256) to model the global dependencies among tokens. The enhanced tokens are then sent to the decoder. Detailed parameters are shown in Table \ref{tbl1}, where the "↓" symbol indicates a downsampling operation.

\begin{table}[width=.9\linewidth,cols=4,pos=h]
\caption{Encoder parameters}
\label{tbl1}
\renewcommand\arraystretch{1.2}
\begin{tabular*}{\tblwidth}{@{\extracolsep{\fill}} c c c c c}
\toprule
\textbf{Layer Type} & \textbf{Size} & \textbf{Repeat} & \textbf{Channels} & \textbf{Stride} \\
\midrule
\centering Image & 224×224 & 1 & 3 & 1 \\
\centering Conv 3×3,↓ & 224×224 & 1 & 16 & 2 \\
\centering Inverted Residual & 112×112 & 1 & 32 & 1 \\
\centering Inverted Residual,↓ & 112×112 & 1 & 64 & 2 \\
\centering Inverted Residual & 56×56 & 2 & 64 & 1 \\
\centering Inverted Residual,↓ & 56×56 & 1 & 128 & 2 \\
\centering MobileViT Block & 28×28 & 2 & 128 & 1 \\
\centering Inverted Residual,↓ & 28×28 & 1 & 256 & 2 \\
\centering MobileViT Block & 14×14 & 4 & 256 & 1 \\
\centering Inverted Residual,↓ & 14×14 & 1 & 320 & 2 \\
\centering MobileViT Block & 7×7 & 3 & 320 & 1 \\
\centering Conv 1×1 & 7×7 & 1 & 256 & 1 \\
\centering Patch Embedding & 49×256 & 1 & 256 & 1 \\
\centering Linear & 49×128 & 1 & 128 & 1 \\
\bottomrule
\end{tabular*}
\end{table}

\subsubsection{TransLPRNet text decoder}\label{}
\begin{figure}[h]%[]
  \centering
    \includegraphics[width=0.8\linewidth]{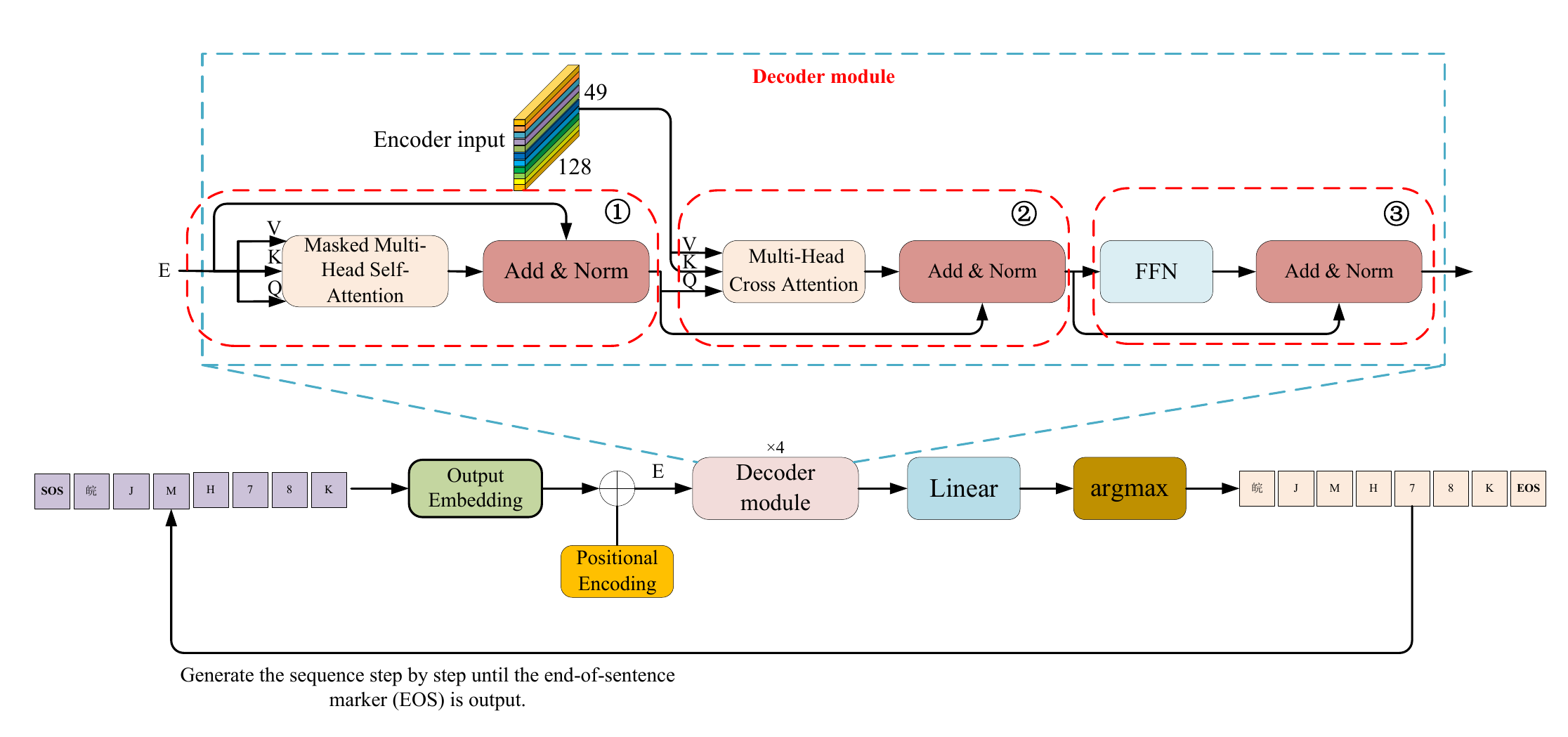}
    \caption{TransLPRNet decoder network structure diagram }\label{fig4}
\end{figure}

As shown in Figure  \ref{fig4}, the core of the text decoder used in the TransLPRNet model consists of four standard Transformer layers. During training, the text decoder first utilizes a masked multi-head self-attention layer to process all known or previously generated target sequence content before the current moment in the target sequence. This masking mechanism ensures that the model does not prematurely see future target information when predicting the current position. Subsequently, through a multi-head cross-attention layer, the decoder fuses the processed target sequence information with the feature sequence output by the visual encoder, thereby using visual information to assist the model in understanding the structure and content of the sequence. This design enables the model to effectively learn the sequential dependencies and contextual constraints in the target sequence, thereby improving the accuracy of recognition and generation.

Specifically, the visual encoder encodes the license plate image into a fixed-length feature sequence ${Z} = \{{{Z}}_1, {{Z}}_2, \ldots, {{Z}}_{49}\}$, where each ${{Z}}_i\in\mathbb{R}^{128}$ represents an image-level semantic token. which serves as the conditional input to the decoder to guide the decoding process to focus on the image content. In the decoder, the hidden state of the current layer after being processed by the masked multi-head self-attention module generates the query Q through a projection matrix W, which is used to guide the decoder to attend to the correlation information within its own sequence. Meanwhile, the encoder output Z generates the key K and value V through independent projection matrices, providing image feature information for the attention computation.

During inference, the input to the decoder consists of two parts: the feature sequence Z output by the visual encoder and the feedback of the historically generated character sequence from the decoder itself. The decoder's inference starts with the initial input start-of-sequence token [SOS]. This [SOS] token serves as the first query token, which is converted into an embedding vector via Embedding Mapping and combined with positional encoding, then used as the input for the first computational step of the decoder. In this step, based on the embedding vector of [SOS], the decoder generates the probability distribution of the first character through Masked Multi-Head Attention (which processes only [SOS] itself at this stage) and Multi-Head Cross Attention (which interacts with the encoder output Z), and then obtains the Token ID of the first character via Argmax. Subsequently, this generated Token ID is appended to the input, together with [SOS], forming the input for the second computational step. The decoder continues to predict the second character based on this longer historical sequence. This process continues iteratively, with the length of the input sequence increasing at each step as generation proceeds, until the [EOS] token is output or the maximum length N=10 (adjustable according to the license plate type) is reached, forming a complete character sequence.

In the license plate recognition task, although the input is an image feature sequence of license plate characters, the self-attention mechanism and position encoding module of the TransLPRNet text decoder possess strong adaptability and can be effectively applied to model the image feature sequence output by the visual encoder. Experimental results show that this strategy effectively alleviates the overfitting problem on small datasets, enabling the lightweight text decoder to achieve high-precision prediction of license plate character sequences while maintaining high efficiency. Table \ref{tbl2} shows the key parameters of the text decoder adopted in this paper.

\begin{table}[width=.9\linewidth,cols=4,pos=h]
\centering
\caption{Network parameters of the TransLPRNet text decoder}
\label{tbl2}
\begin{tabular*}{\tblwidth}{@{\extracolsep{\fill}} c c }
\toprule
Parameter Name & Number of Parameters \\
\midrule
Model Layers & 4 \\
Hidden Dimensions & 128 \\
Number of attention heads & 4 \\
FFN hidden layer dimension & 512 \\
\bottomrule
\end{tabular*}
\end{table}

\subsection{  Perspective Transformation Network: PTN}\label{}
The widely utilized Spatial Transformer Network (STN) \cite{jaderberg2015spatial} commonly performs image rectification by estimating a 6-parameter affine transformation matrix. Although theoretically extendable to an 8-parameter model for perspective transformations, direct application of STN for license plate image perspective rectification in practice often yields noticeably distorted and unnatural results. This is primarily due to the high difficulty in parameter regression, training instability, and a lack of effective direct supervision (as detailed in Section 5.2).

To address these challenges, this paper proposes a novel Perspective Transformation Network (PTN). Instead of directly regressing the transformation matrix as in STN, PTN combines the estimation of license plate corner coordinates with the explicit computation of the perspective transformation matrix. This approach not only effectively resolves the difficulties STN faces in accurately estimating spatial transformation parameters for perspective-distorted license plates, but also offers enhanced interpretability, as the estimated corner coordinates can be visualized as needed.

Similar to STN, the proposed PTN could, in principle, learn from supervisory signals provided by a downstream license plate recognition network. However, given that our chosen recognition network, transLPRNet, inherently possesses some robustness to variations in capture angle, using its feedback to supervise PTN, while improving recognition accuracy, often leads to rectification results that are visually inconsistent with human perception. Therefore, we introduce an innovative training scheme: leveraging weak supervision provided by a dedicated license plate view classification network to train PTN. This approach significantly simplifies data annotation (where only front-view license plate images are labeled as positive, and others as negative) and effectively decouples PTN from the recognition network. Consequently, transLPRNet can be trained with diverse angle license plate images, while during inference, it only needs to process images rectified by PTN, thereby substantially enhancing recognition accuracy by mitigating the interference caused by varying capture angles.

At its core, PTN transforms the traditional task of regressing spatial transformation matrix parameters into the regression of the four corner coordinates (e.g., (x1,y1)...(x4,y4)) of the license plate region. Specifically, PTN comprises a license plate corner coordinate regression sub-network responsible for estimating these four vertices from the input image. These estimated coordinates are then fed into a perspective transformation matrix computation module, which, utilizing the inverse perspective transform formula, derives the complete transformation parameters required to rectify the license plate from its current distorted pose to a canonical front view. The overall architecture of PTN is illustrated in Figure 5; its grid generation and sampling modules are adopted from the original STN design.
\begin{figure}%[]
  \centering
    \includegraphics[width=0.8\linewidth]{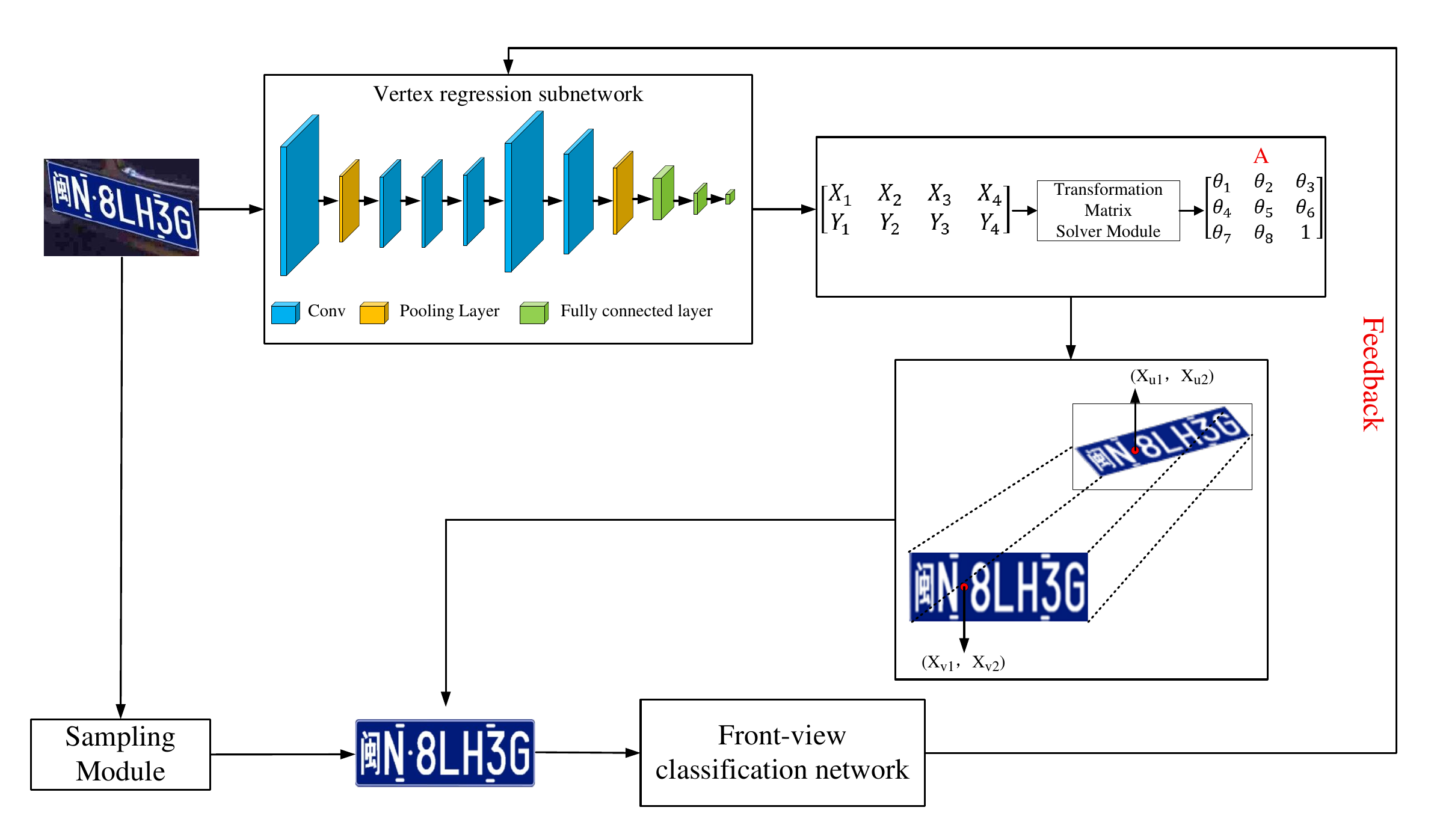}
    \caption{PTN correction network system block diagram}\label{fig5}
\end{figure}

\subsubsection{ License Plate Vertex Coordinate Regression Sub-network}\label{}
This sub-network is designed to perform the coordinate regression of license plate vertices, with an input image size of 94×24 pixels. The network first extracts features from the license plate region by employing a multi-layer convolutional structure to obtain rich local feature information. 

Subsequently, two pooling layers are utilized to progressively reduce the spatial dimensions of the feature maps, thereby enhancing the abstract representation capability of the features. In the deep feature extraction phase, the network merges and integrates the high-level features through three fully connected layers, ultimately achieving the regression of the license plate vertex coordinates.Specifically, this sub-network comprises 7 convolutional layers, 2 pooling layers, and 3 fully connected layers. The convolutional layers are responsible for local spatial feature extraction, while the pooling layers reduce the spatial dimensions of the features to enhance the model's robustness to noise. The final three fully connected layers map the extracted high-level features to eight parameters, which correspond to the two-dimensional coordinates of the four vertices of the license plate $({X}_i, {Y}_i)$,where $i = 1, 2, 3, 4$.

\begin{figure}[h]
  \centering
  \includegraphics[width=0.8\linewidth]{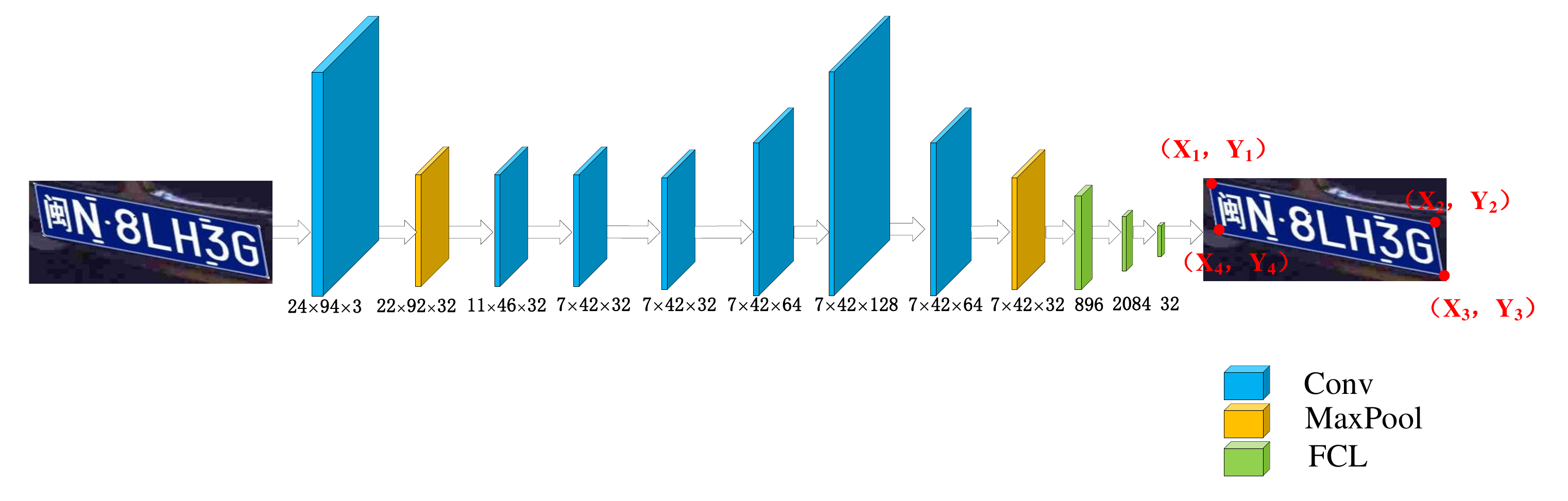}
  \caption{Vertex regression subnetwork}
  \label{fig6}
\end{figure}

Figure \ref{fig6} illustrates the architecture of this sub-network. The input is a cropped license plate region image, and through forward propagation, the network accurately regresses the spatial positions of the four license plate vertices. This provides essential geometric information for subsequent tasks such as geometric correction and license plate recognition.Table \ref{tb99}  provides a detailed overview of the network architecture and parameters of the license plate vertex regression subnet.

\begin{table}[width=.9\linewidth,cols=4,pos=h]
\caption{Architecture and parameters of the license plate vertex regression network}\label{tb99}
\begin{tabular*}{\tblwidth}{@{}CCCCC@{} }
\toprule
Name&Convolution kernel size&Convolution stride&Input size&Output size\\
\midrule
Conv&3×3&-&24×94×3&22×92×32\\
MaxPool&2×2&2&22×92×32&11×46×32\\
Leaky ReLU&-&-&11×46×32&11×46×32\\
Conv&5×5&-&11×46×32&7×42×32\\
Conv&3×3&1&7×42×32&7×42×32\\
Conv&3×3&1&7×42×32&7×42×64\\
Leaky ReLU&-&-&7×42×64&7×42×64\\
Conv&3×3&1&7×42×64&7×42×128\\
Conv&1×1&-&7×42×128&7×42×64\\
Leaky ReLU&-&-&7×42×64&7×42×64\\
Conv&3×3&1&7×42×64&7×42×32\\
MaxPool&3×3&3&7×42×32&2×14×32\\
Leaky ReLU&-&-&2×14×32&2×14×32\\
FCL&-&-&896&2084\\
FCL&-&-&2084&32\\
FCL&-&-&32&8\\
\bottomrule
\end{tabular*}
\end{table}

\subsubsection{ Perspective Transformation Matrix Estimation Module}\label{}
This module computes the perspective transformation matrix based on the four vertices of the license plate and the four corner points of the output corrected image.By using the perspective inverse transformation formula \cite{zhang1999flexible}, the parameters can ultimately be calculated by solving the following linear equations: 
\begin{equation}
\begin{bmatrix}
   X_{n1} & Y_{n1} & 1 & 0 & 0 & 0 & -U_{m1}X_{n1} & -U_{m1}Y_{n1} \\ 
   0 & 0 & 0 & X_{n1} & Y_{n1} & 1 & -V_{m1}X_{n1} & -V_{m1}Y_{n1} \\
   X_{n2} & Y_{n2} & 1 & 0 & 0 & 0 & -U_{m2}X_{n2} & -U_{m2}Y_{n2} \\ 
   0 & 0 & 0 & X_{n2} & Y_{n2} & 1 & -V_{m2}X_{n2} & -V_{m2}Y_{n2} \\
   X_{n3} & Y_{n3} & 1 & 0 & 0 & 0 & -U_{m3}X_{n3} & -U_{m3}Y_{n3} \\ 
   0 & 0 & 0 & X_{n3} & Y_{n3} & 1 & -V_{m3}X_{n3} & -V_{m3}Y_{n3} \\
   X_{n4} & Y_{n4} & 1 & 0 & 0 & 0 & -U_{m4}X_{n4} & -U_{m4}Y_{n4} \\ 
   0 & 0 & 0 & X_{n4} & Y_{n4} & 1 & -V_{m4}X_{n4} & -V_{m4}Y_{n4} \\
\end{bmatrix}
\begin{bmatrix}
   \theta_1 \\  
   \theta_4 \\ 
   \theta_7 \\ 
   \theta_2 \\
   \theta_5 \\
   \theta_8 \\
   \theta_3 \\
   \theta_6 \\
\end{bmatrix}
=
\begin{bmatrix}
   U_{m1} \\  
   V_{m1} \\ 
   U_{m2} \\  
   V_{m2} \\
   U_{m3} \\  
   V_{m3} \\
   U_{m4} \\  
   V_{m4} \\
\end{bmatrix}
\end{equation}

\begin{figure}[htbp]
  \centering
  \includegraphics[width=0.6\linewidth]{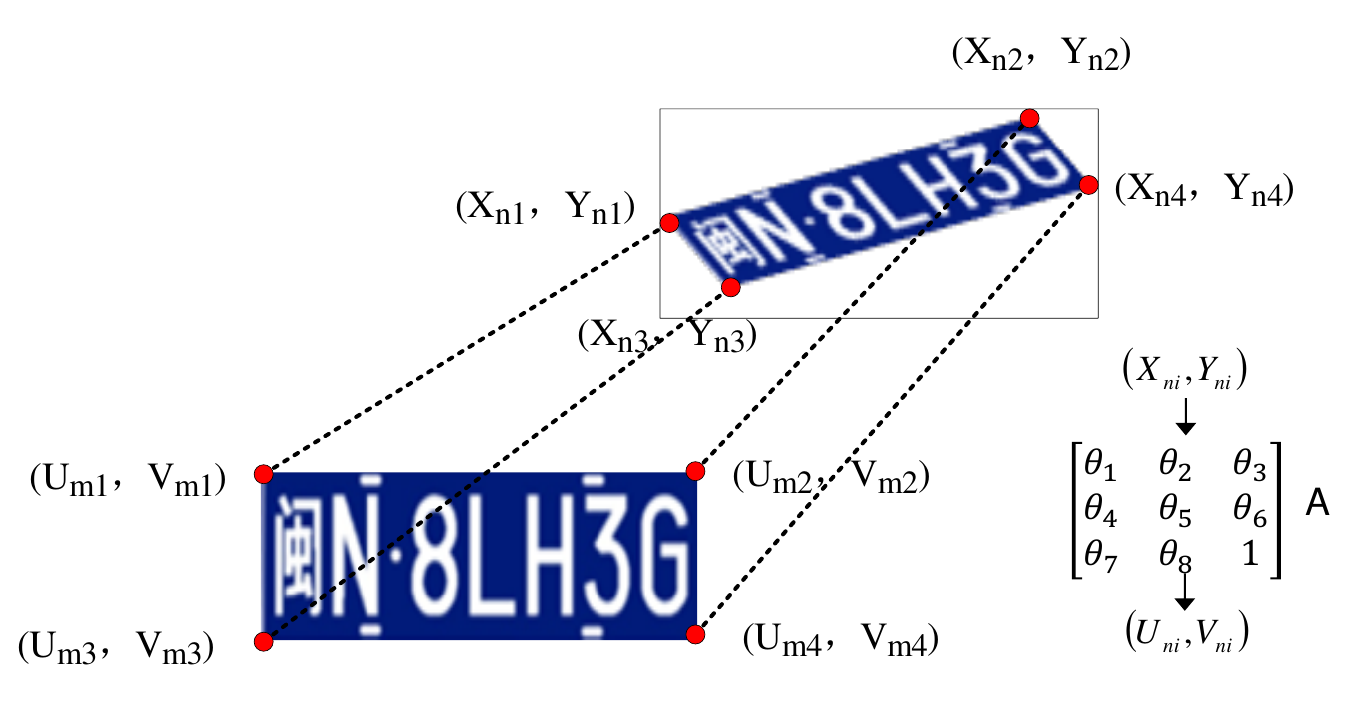}
  \caption{Mapping diagram between the LP four vertices and the four corner points of the input or output image}
  \label{fig7}
\end{figure}
Here $({U}_{mi}, {V}_{mi})$($i = 1, 2, 3, 4$) denote the coordinates of the four target points—top-left, top-right, bottom-left, and bottom-right—on the output license plate image. Meanwhile, $({X}_{ni}, {Y}_{ni})$($i = 1, 2, 3, 4$) represent the coordinates of the corresponding source points within the input license plate image, which also serve as the four corner points of the desired rectified image. By solving the resulting system of linear equations (as shown in formula 2), eight parameters, namely $\theta_1, \theta_2,\theta_3, \theta_4, \theta_5, \theta_6, \theta_7, \theta_8$ are obtained. These parameters constitute the first eight elements of the perspective inverse transform matrix A in figure \ref{fig5}. The ninth parameter, typically set to 1 as a scaling factor, is appended to complete the matrix. Consequently, the nine parameters are assembled into a 3×3 perspective transformation matrix A. The mapping relationship between the four vertices of the license plate and the four corner points of either the original image or the rectified output image is illustrated in Figure \ref{fig7}.

\subsubsection{Classifier-Guided Weak Supervision for PTN in License Plate Rectification}\label{}
\begin{figure}[htbp]
  \centering
  \includegraphics[width=0.8\linewidth]{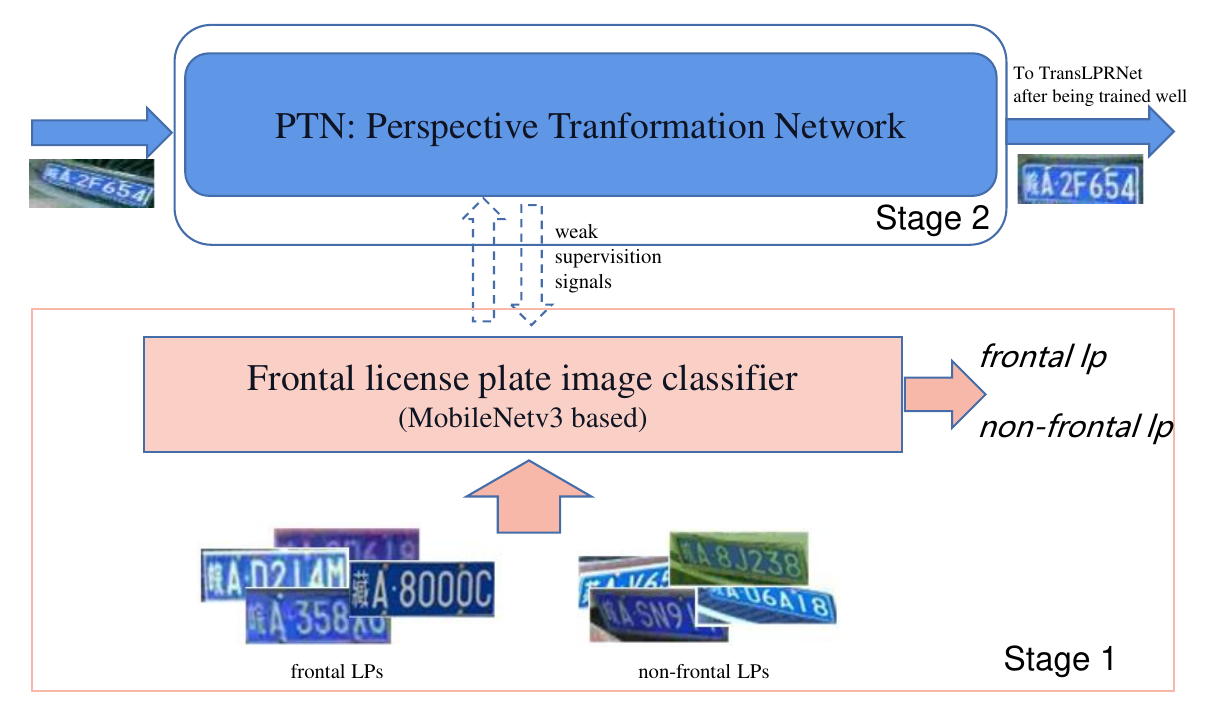}
  \caption{illustration of PTN supervision signals from the frontal license plate image classifier}
  \label{fig8}
\end{figure}

To provide PTN with an independent supervision signal decoupled from high-level recognition networks, this paper introduces a lightweight MobileNetV3-based \cite{howard2019searching} frontal license plate classifier after the PTN. This classifier performs binary classification on license plate images processed by PTN, directly indicating whether the input image has been successfully rectified into a standard frontal view. This classification result then serves as crucial feedback for PTN. Considering that the feedback from this classifier is a weak supervision signal relative to PTN's detailed geometric rectification task, we propose a two-stage training method. First, the MobileNetV3 frontal license plate classifier is independently trained using annotated frontal and non-frontal license plate images (as shown in Stage 1 of Figure \ref{fig8}). Subsequently, after freezing the parameters of this classifier, it is integrated with the PTN model to enable end-to-end training of the PTN (as shown in Stage 2 of Figure \ref{fig8}). Detailed experimental setups and result analysis will be elaborated in the experimental section of this paper.

\subsection{Construction of a Single/Double-Line License Plate Image Dataset}\label{}
License plate image datasets collected in unconstrained environments should exhibit sample diversity, encompassing a wide range of angles, lighting conditions, and various interferences to more accurately reflect the complexities of real-world usage. Several publicly available license plate datasets are currently employed in research, such as CLPD \cite{zhang2020robust} and LSV-LP \cite{wang2022lsv}. While these datasets demonstrate certain effectiveness in specific scenarios, they often suffer from limitations like insufficient environmental diversity and a limited number of license plate samples.In contrast, the CCPD \cite{xu2018towards} dataset is more adaptable for license plate recognition in unconstrained environments. As a publicly available dataset specifically designed for Chinese license plates, it comprises seven subsets with approximately 280,000 blue license plate images. These images cover a variety of scenes, angles, lighting conditions, and weather. The dataset provides detailed annotation information, including license plate locations and numbers.

Despite being the most widely used public dataset for research in license plate recognition in unconstrained environments, the CCPD dataset still has inherent issues. First, its annotation employs an iterative strategy, where a model is trained on a portion of the annotated data, then used to predict the vertices and bounding boxes of the license plates. This process can introduce annotation errors. Second, the dataset only contains single-line license plates, lacking samples of double-line plates. Furthermore, although the subsets cover various scenes, the distribution of license plate samples across different scenes is imbalanced. Some scenes have abundant samples, while others are severely lacking, which limits its potential applications.To address these problems, this paper proposes two improvements: First, we correct the mislabeled license plates. Second, we introduce a strategy to integrate multi-type license plates (single-line and double-line), by pasting synthetic double-line license plates onto redundant single-line license plate images; thereby building a more diverse and comprehensive license plate dataset, which enables improving robustness and generalization capabilities for LPR algorithms.
\subsubsection{Preprocessing of License Plate Image Dataset Used in the Experiments}\label{}
\begin{figure}[htbp]
  \centering
  \includegraphics[width=0.6\linewidth]{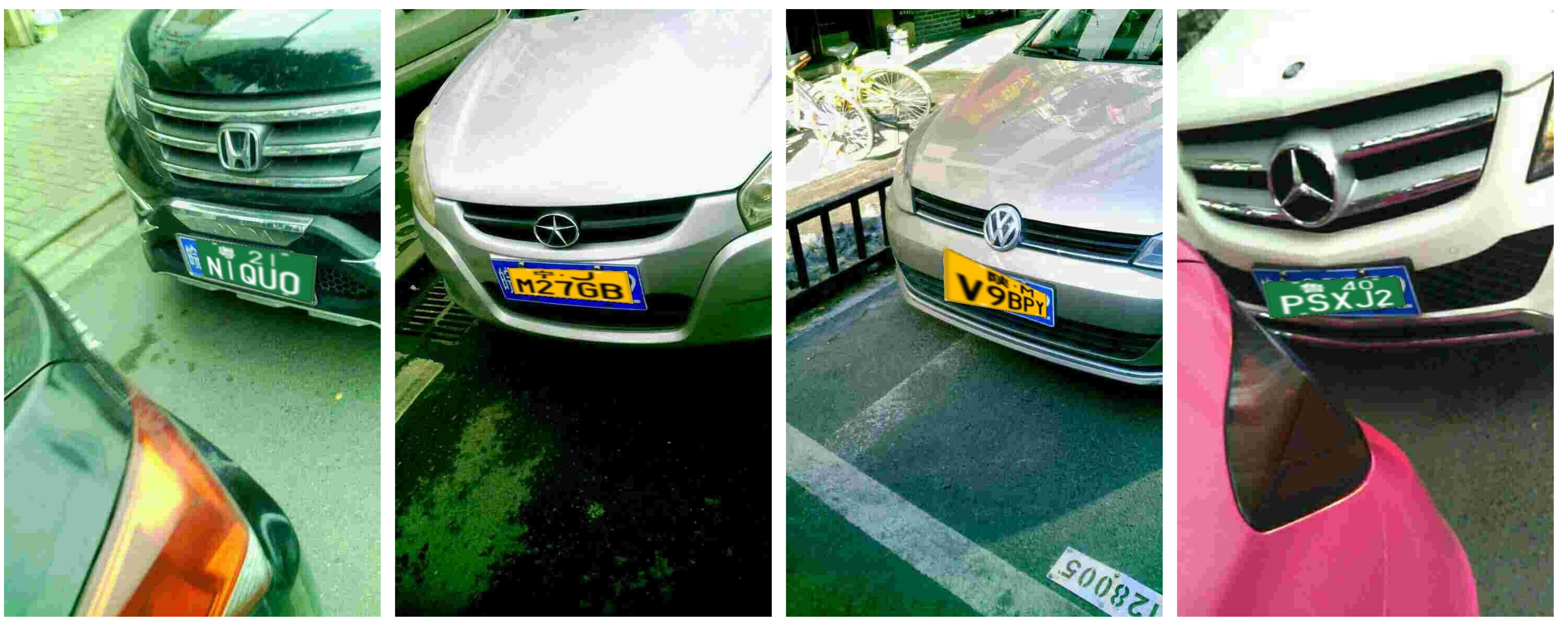}
  \caption{Illustration of license plate images before label correction }
  \label{fig9}
\end{figure}

\begin{figure}[htbp]
  \centering
  \includegraphics[width=0.6\linewidth]{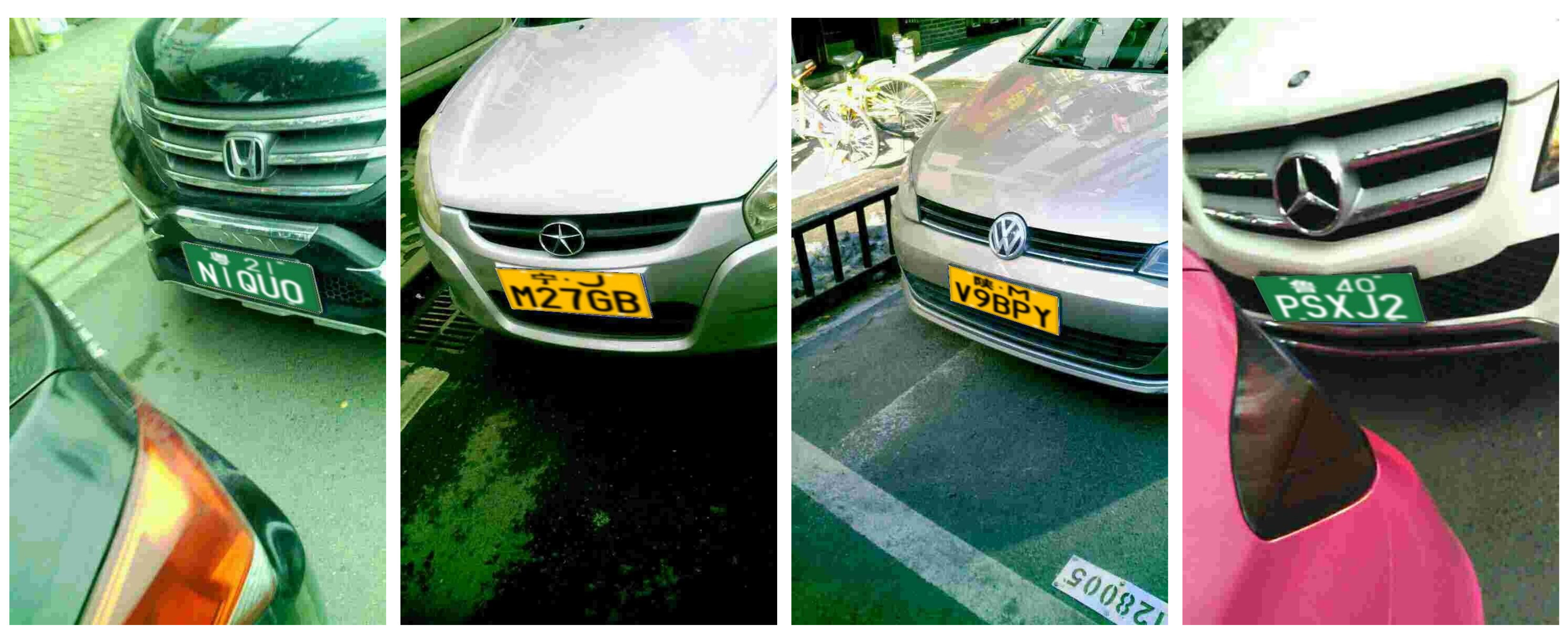}
  \caption{Illustration of license plate images after label correction }
  \label{fig10}
\end{figure}

This paper focuses on license plate image correction and recognition, and does not involve the localization process. To evaluate the impact of real-world license plate localization errors on subsequent correction and recognition performance, we utilize the bounding boxes and four-vertex coordinate information provided by the CCPD dataset, and apply random perturbations on top to simulate the potential effects of actual localization inaccuracies on the license plate images.During the experiments, we found that the original CCPD dataset contains annotation errors, mainly because its labels are generated through a combination of manual annotation and model predictions, which can inevitably introduce inaccuracies. As shown in Figure \ref{fig9}, mapping images using the original CCPD labels results in noticeable misalignments in some cases; in contrast, Figure \ref{fig10} illustrates that after our systematic correction, the label mappings are more accurate and properly aligned.Therefore, directly using uncorrected labels for image mapping and applying random disturbances could significantly affect the accuracy of subsequent license plate correction and recognition. To ensure the reliability and validity of the experimental results, we performed a systematic correction of the label errors in the CCPD dataset before conducting further experiments.

\begin{figure}[htbp]
  \centering
  \includegraphics[width=0.6\linewidth]{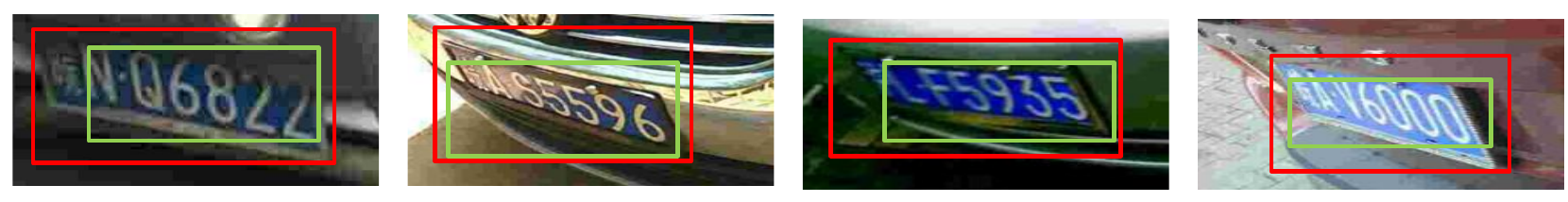}
  \caption{Illustration of detection results versus original labels, where the red boxes indicate the model's predicted license plate locations, and the green boxes represent the original annotated license plate locations}
  \label{fig11}
\end{figure}

This paper employs the license plate detection model used in the CCPD study \cite{xu2018towards} to detect license plates in the images. Subsequently, the Intersection Over Union (IOU) between the predicted bounding boxes and the annotated ground-truth bounding boxes is calculated. If the IOU exceeds 0.6, the annotation is considered correct; otherwise, it is deemed to contain an error. As illustrated in Figure \ref{fig11}, among the filtered samples identified as having incorrect annotations, the red boxes indicate the model's predicted license plate locations, while the green boxes represent the original ground-truth annotations. Through this filtering process, a total of 1,414 images with annotation errors were identified. For these problematic samples, manual re-annotation was performed to correct their labels, resulting in a refined, accurately annotated license plate dataset that provides a more reliable foundation for subsequent model training.
\subsubsection{Double/Single-Line License Plate Dataset Construction via Image Overlay}\label{}

To address the issue that the insufficient dataset of double-line license plates severely restricts the training performance of the TransLPRNet model, this paper adopts an image pasting method. While preserving the original single-line license plate images from CCPD, this approach effectively compresses and utilizes existing data resources while increasing the diversity of double-line license plate samples.

Specifically, the Base subset of the CCPD dataset was first randomly and evenly divided into two parts: one part containing 100,000 images for model training, and the other part containing another 100,000 images for model testing. Given that the original authors of CCPD adopted the RPNet\cite{xu2018towards} model for license plate detection and recognition in their paper, this study selected the recognition model from RPNet as the baseline model for training. The trained baseline model was then used to recognize images in the test set, and those images that were correctly recognized were selected. Since these images could be correctly recognized by the model trained on the training set, they were considered redundant samples relative to both the training and test sets. Therefore, 50,000 license plate images were selected from these redundant samples. Using their real-world background and shooting angle information, 25,000 double-line yellow license plates and 25,000 double-line green license plates were constructed. Among these, 25,000 double-line plates were added to the training set, and the corresponding single-line real license plate images were randomly replaced and moved into the test set. The remaining 25,000 double-line license plates were used as the Base-d subset within the test set to evaluate the model's recognition performance on double-line license plate images.

Although the above processing results in the loss of 25,000 single-line license plate images from the training set, the impact on training is minimal, as these license plates satisfy the independent and identically distributed condition and are randomly sampled. For the test set, since only redundant samples are used for image pasting, the impact on test results is also very small. Although the RPNet model differs from the model used in this paper, the difficult-to-recognize samples share common characteristics across different models. This approach not only preserves the information of single-line license plates to the greatest extent, but also increases the difficulty of both the training and test sets for single-line plates to a certain degree, without increasing the size of the dataset. Throughout the entire process, there is no data leakage between the training and test sets.

\begin{figure}[htbp]
  \centering
  \includegraphics[width=0.6\linewidth]{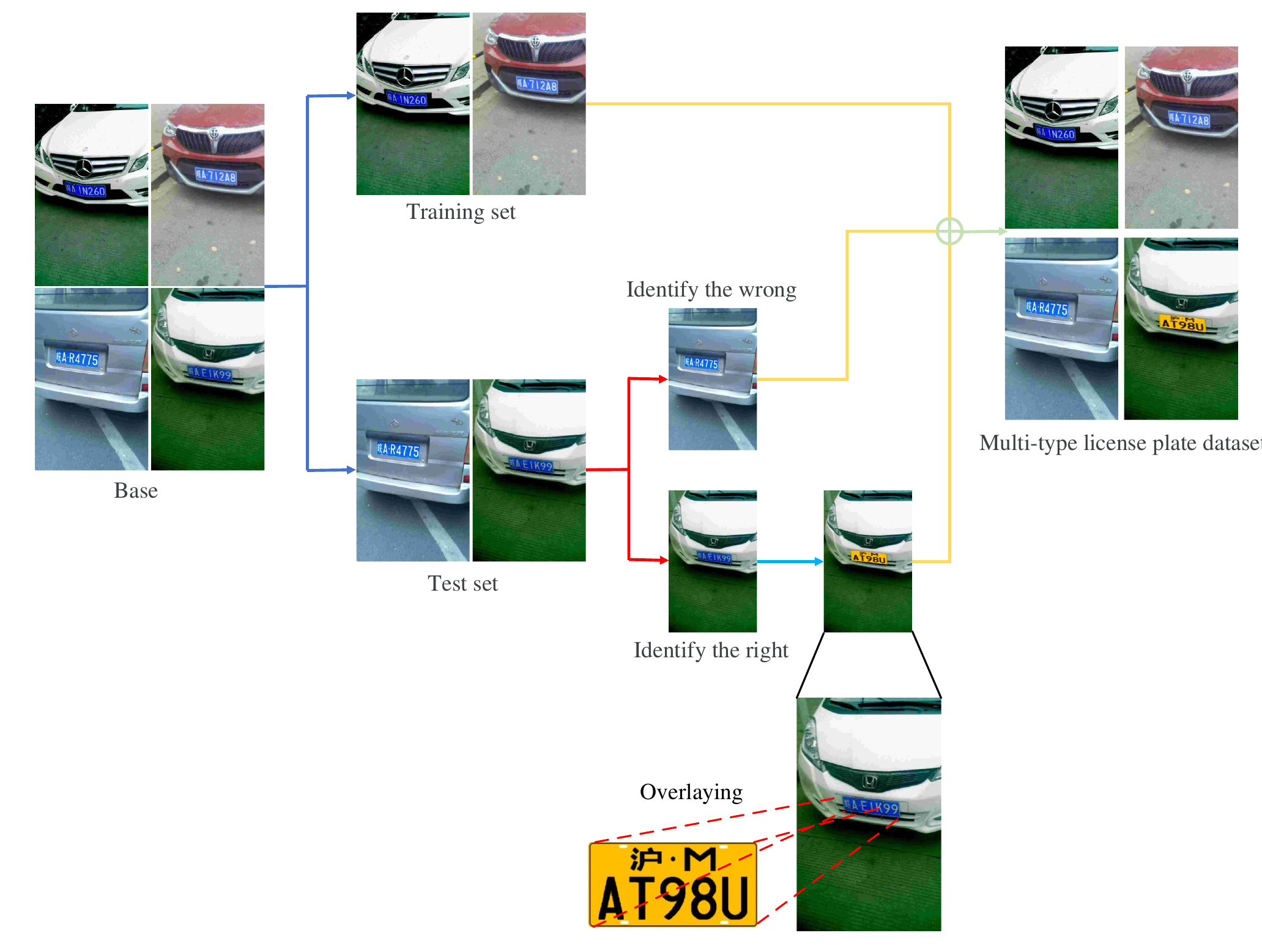}
  \caption{Flowchart of composite training dataset construction}
  \label{fig12}
\end{figure}

\begin{figure}[htbp]
  \centering
  \includegraphics[width=0.8\linewidth]{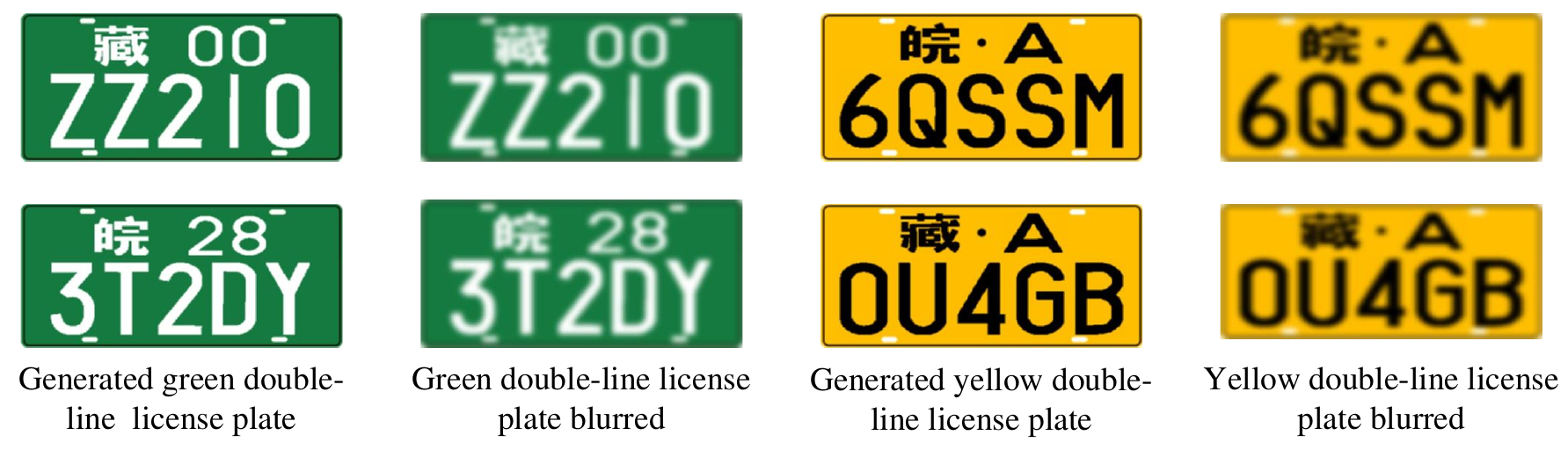}
  \caption{Schematic diagram of double-line license plates and blurring}
  \label{fig13}
\end{figure}

\begin{figure}[htbp]
  \centering
  \includegraphics[width=0.6\linewidth]{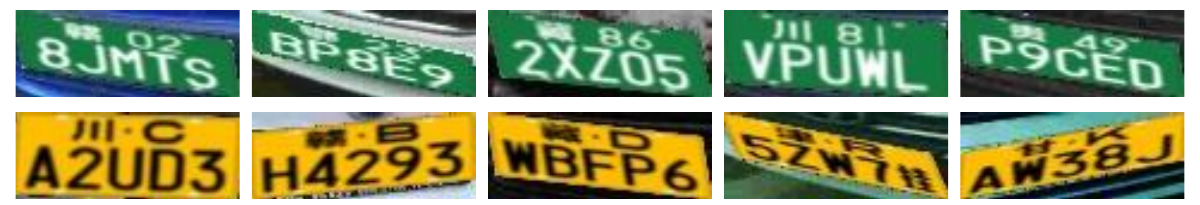}
  \caption{Example of cropped images from the double-line license plate test set}
  \label{fig14}
\end{figure}

The specific construction process of the double-line license plate dataset based on image pasting is illustrated in Figure \ref{fig12}. First, the program-generated double-line license plate images are blurred to produce blurred samples of yellow and green double-line license plates (as shown in Figure \ref{fig13}), which are used to simulate image blur and environmental changes in actual captures. Next, using image pasting techniques, these synthesized double-line license plate images are overlaid onto redundant license plate samples via perspective transformation. Finally, based on the annotation information of the bounding boxes (after random perturbation), the license plate regions are cropped out (as shown in Figure \ref{fig14}), forming a mixed dataset containing both single-line and double-line license plates.

\begin{figure}[htbp]
  \centering
  \includegraphics[width=0.6\linewidth]{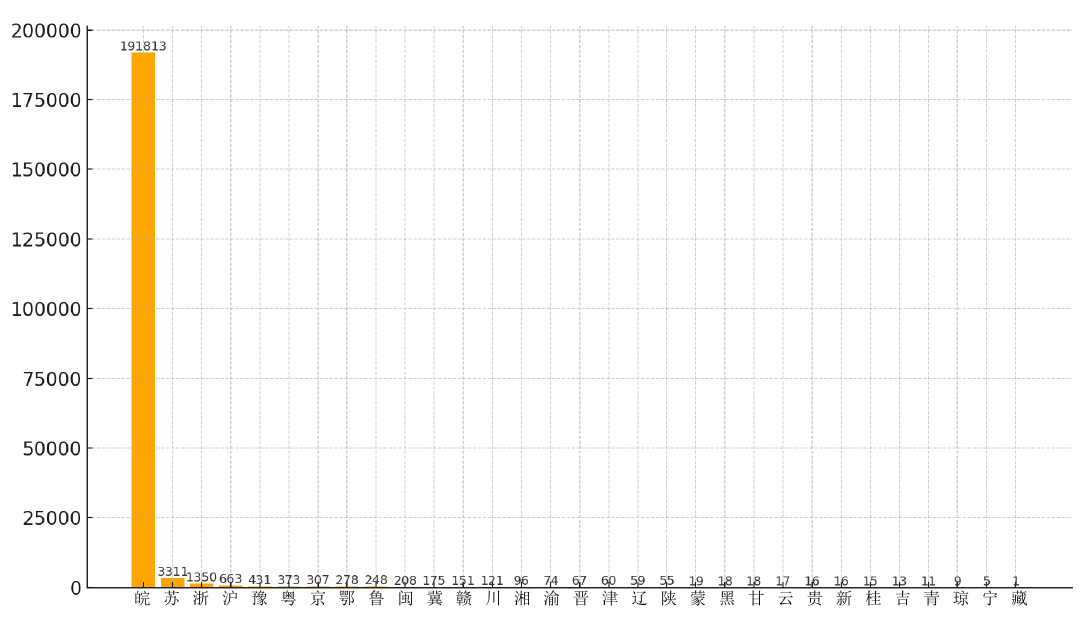}
  \caption{Histogram of provincial distribution before dataset augmentation}
  \label{fig15}
\end{figure}

\begin{figure}[htbp]
  \centering
  \includegraphics[width=0.6\linewidth]{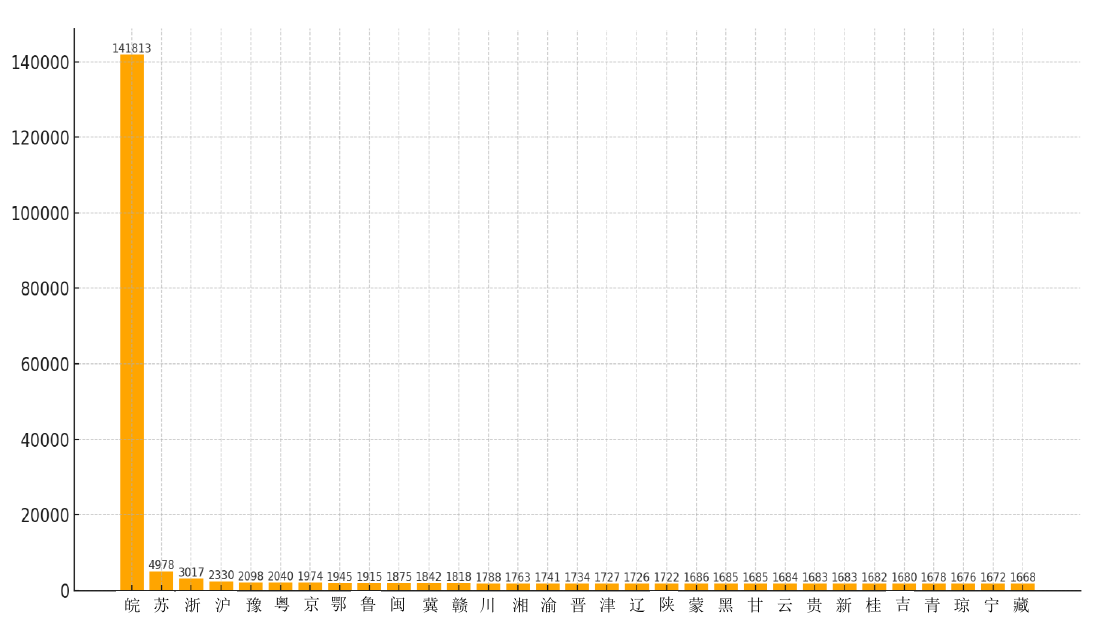}
  \caption{Histogram of provincial distribution after dataset augmentation}
  \label{fig16}
\end{figure}

\begin{figure}[htbp]
  \centering
  \includegraphics[width=0.6\linewidth]{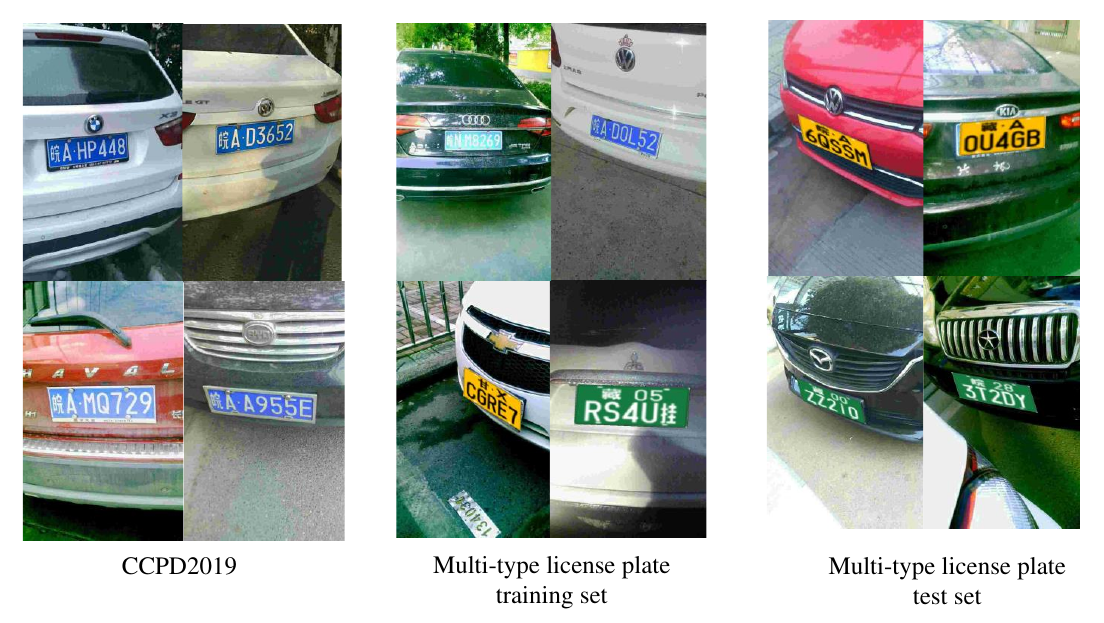}
  \caption{Example of overlaying generated double-line license plates onto CCPD dataset images}
  \label{fig17}
\end{figure}

\begin{figure}[htbp]
  \centering
  \includegraphics[width=0.6\linewidth]{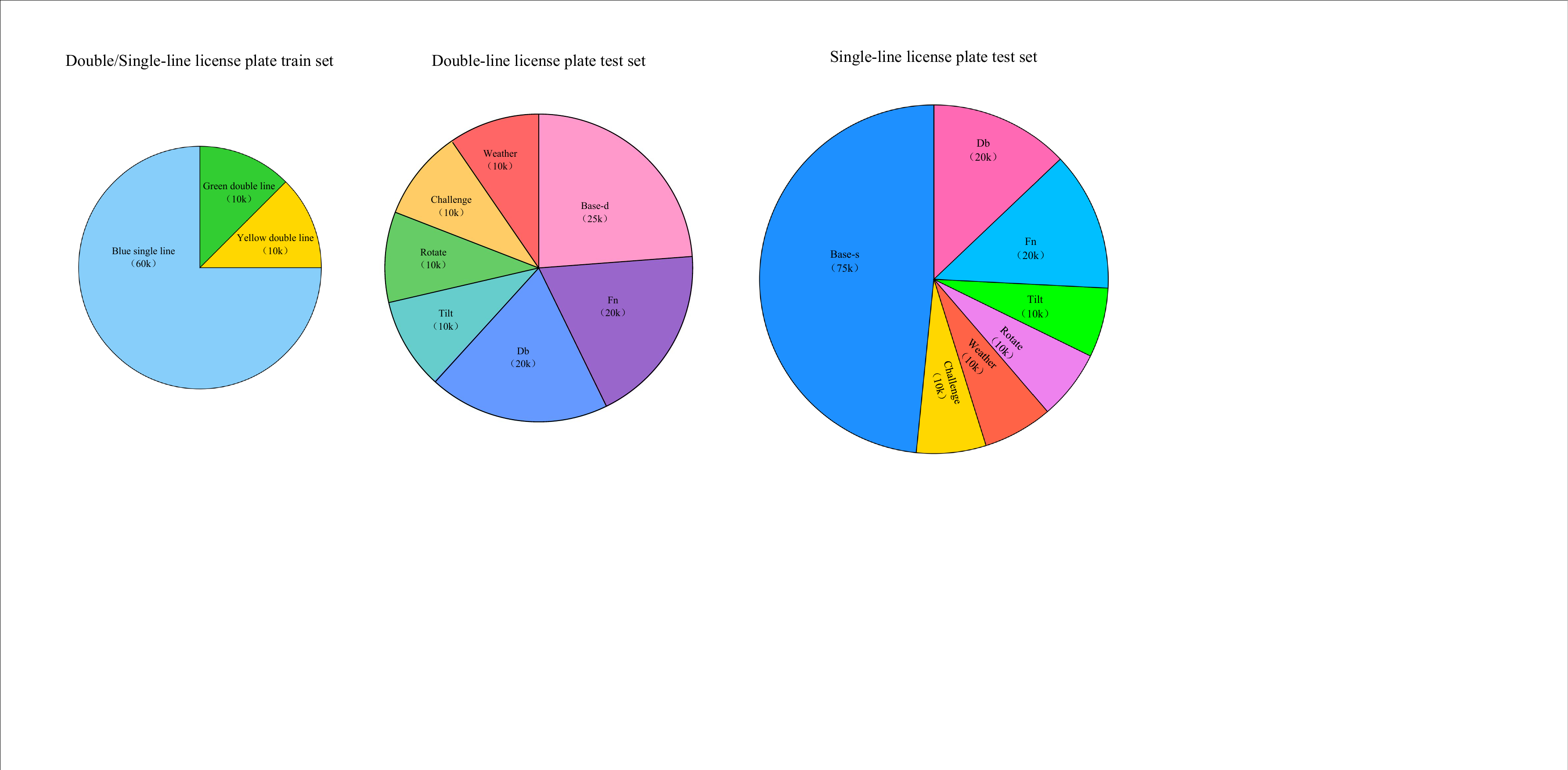}
  \caption{Pie chart of the distribution of the single/double-line license plate training set and single/double-line license plate test set}
  \label{fig18}
\end{figure}

Figure \ref{fig15} shows a histogram of the distribution of license plates by province in the Base dataset before expansion, while Figure \ref{fig16} shows the histogram of the provincial distribution of license plates in the Base dataset after expansion. To ensure the diversity of data augmentation and the generalization ability of the recognition model, the double-line license plate images used for data enhancement are kept uniform in both quantity proportion and provincial distribution, effectively avoiding the risk of the model overfitting to high-frequency provinces such as "Anhui" (Wan). Ultimately, this paper forms a mixed CCPD dataset as shown in Figure \ref{fig17}. The pie chart showing the distribution of the training set and test set is presented in Figure \ref{fig18}.


\section{Experiment settings and result analysis}\label{}
The training environment is as follows: Ubuntu 18.04, PyTorch 1.8, CUDA 11.7, cuDNN 8.5, GTX TITAN X (GPU). The testing environment is as follows: Ubuntu 22.04, PyTorch 2.3, CUDA 11.8, cuDNN 8.9, GTX 1080 Ti (GPU).
\subsection{Simulation of License Plate Detection Errors Based on Random Coordinate Perturbations}\label{}
\begin{figure}[htbp]
  \centering
  \includegraphics[width=0.6\linewidth]{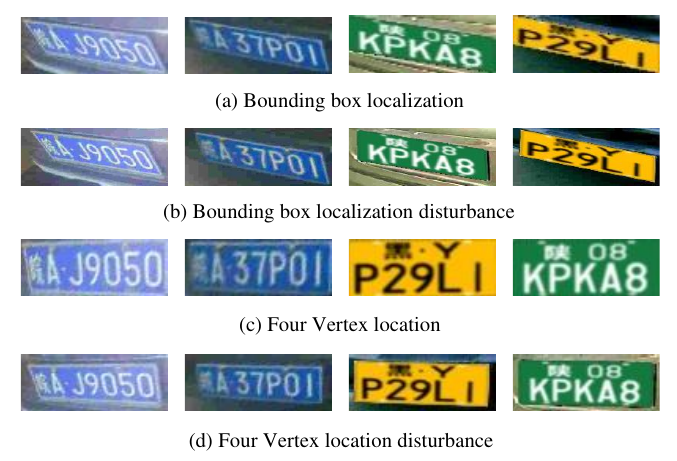}
  \caption{Illustration of coarse and fine perturbations in license plate localization}
  \label{fig19}
\end{figure}

To simulate potential errors inherent in license plate detection, we introduce random coordinate perturbations to meticulously calibrated license plate image labels. This perturbation method involves adding a Gaussian-distributed random offset, characterized by a mean of 0 and a standard deviation of 4, to the original label coordinates. Specifically, for bounding box-based detection, the coordinates of the top-left and bottom-right corner points are independently perturbed(Bounding box localization disturbance). In contrast, for fine-grained localization utilizing license plate vertices, random deviations are applied to the coordinates of all four corner points(Four Vertex location disturbance). This approach aims to effectively mimic realistic error scenarios encountered during practical license plate detection, as visually demonstrated in Figure \ref{fig19}, and enhances the robustness and generalization capabilities of the model.
\subsection{PTN Training and Evaluation }\label{}
\begin{figure}[htbp]
  \centering
  \includegraphics[width=0.6\linewidth]{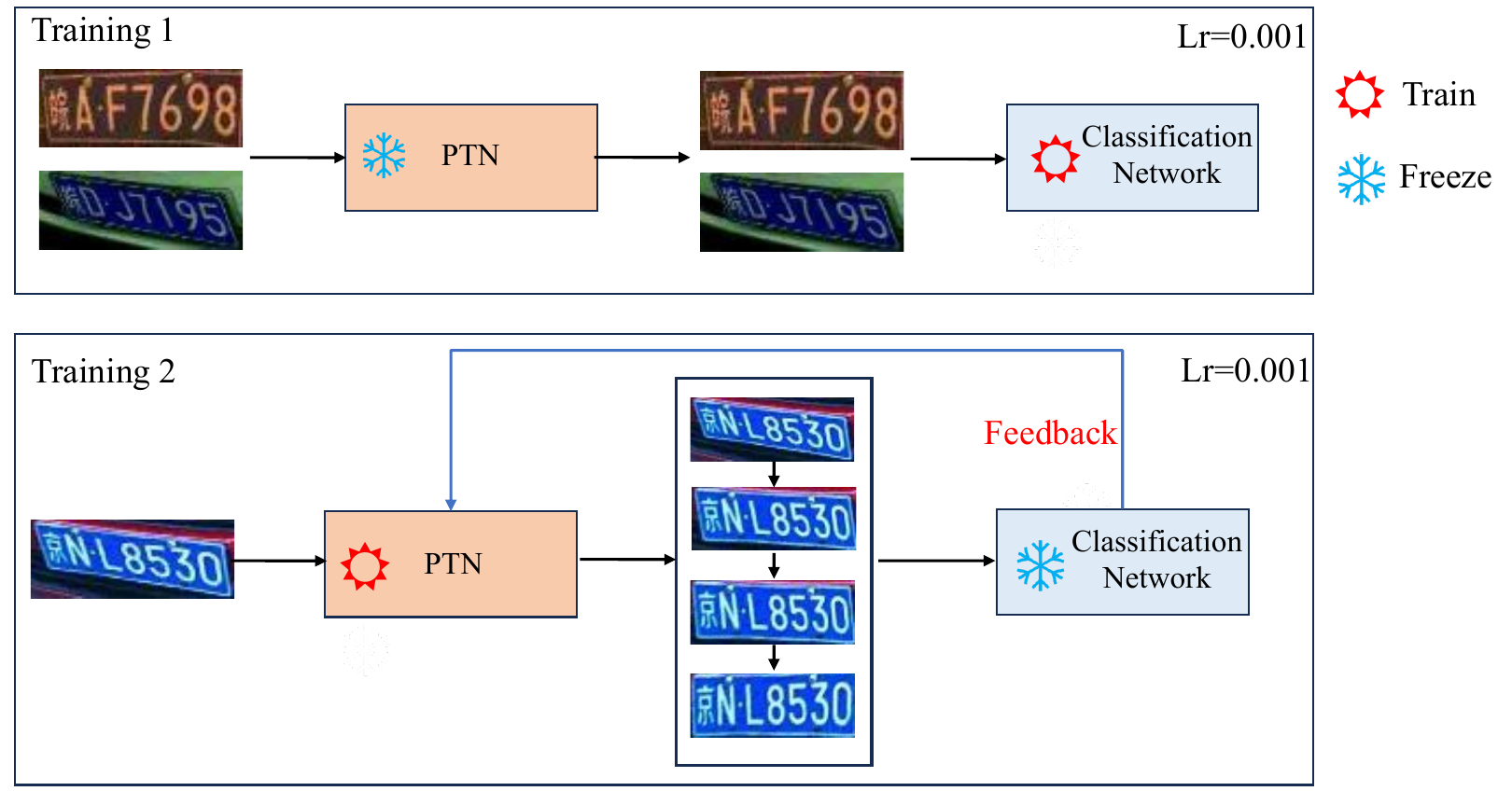}
  \caption{Training workflow of the PTN}
  \label{fig20}
\end{figure}

To further enhance the convergence stability of the license plate correction module based on weakly supervised classification information during training, a two-stage training strategy was employed. In the first stage, the correction network (PTN) was frozen, and only the classification network was trained to ensure robust feature extraction capabilities. In the second stage, the classification network was frozen, and the correction network (PTN) was trained using a randomized mixture of positive and negative samples to prevent alterations in feature extraction patterns.Throughout the training process, a relatively low learning rate was adopted to enable fine-grained adjustments of the network parameters, thereby effectively improving overall model performance. The hyperparameter configuration used in this experiment is detailed in Table \ref{tbl3}, and the training process is illustrated in Figure \ref{fig20}.

\begin{table}[width=.9\linewidth,cols=4,pos=h]
\caption{Hyperparameter configuration for PTN training}\label{tbl3}
\begin{tabular*}{\tblwidth}{@{} CC@{} }
\toprule
\textbf{Parameter Name} & \textbf{Number of Parameters} \\
\midrule
Learning Rate & 0.001 \\
Optimizer & Adam \\
Batch Size & 32 \\
Training Epochs & 100 \\
Image Size & 94×24 \\
\bottomrule
\end{tabular*}
\end{table}

\begin{figure}[htbp]
  \centering
  \includegraphics[width=0.6\linewidth]{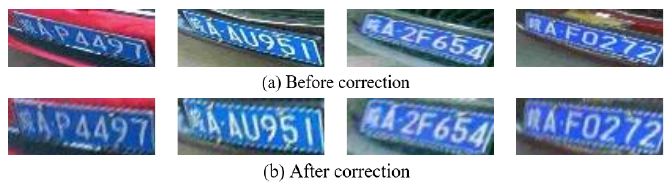}
  \caption{Example of STN before and after affine transformation correction }
  \label{fig21}
\end{figure}

\begin{figure}[htbp]
  \centering
  \includegraphics[width=0.6\linewidth]{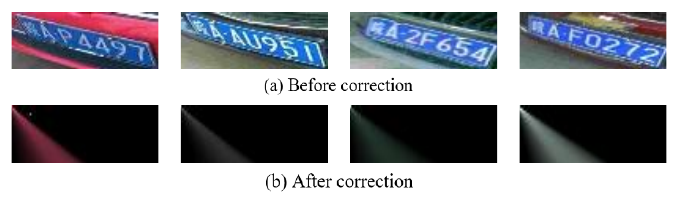}
  \caption{Example images before and after STN perspective transformation correction }
  \label{fig22}
\end{figure}

\begin{figure}[htbp]
  \centering
  \includegraphics[width=0.6\linewidth]{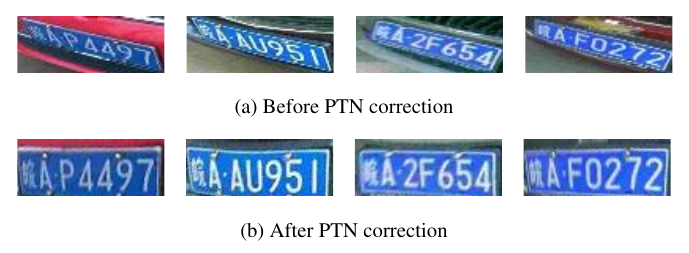}
  \caption{Comparison of images before and after PTN correction }
  \label{fig23}
\end{figure}

In theory, the Spatial Transformer Network (STN) \cite{jaderberg2015spatial} is capable of learning parameters for the spatial geometric transformation of images. However, the STN is primarily designed for spatial correction related to affine transformations, such as image translation, rotation, and scaling. Directly employing an STN to regress a perspective transformation matrix often leads to convergence challenges during network training. This paper presents a preliminary investigation into using STNs to regress both a 6-parameter affine transformation matrix and an 8-parameter perspective transformation matrix. Results of these experiments are shown in Figures \ref{fig21} and \ref{fig22}. The experimental results indicate that the STN provides limited correction of car license plate tilt when regressing affine transformations. Conversely, when the STN is used to regress perspective transformation parameters, the output images exhibit large areas of black, rendering complete license plate content unrecognizable. This phenomenon is primarily attributable to the high degree of interdependence between parameters in the perspective transformation matrix. Slight variations in any parameter within the perspective transformation can induce complex and non-linear geometric distortions in the output image, leading to significant stretching or compression of the license plate image, and ultimately resulting in substantial invalid regions in the transformed image.Figure \ref{fig23} shows a comparison of images before and after PTN correction proposed in this paper.


\begin{figure}[htbp]
  \centering
  \includegraphics[width=0.6\linewidth]{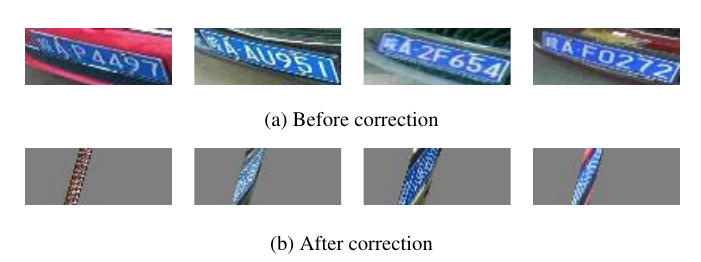}
  \caption{Experimental results of PTN integrated with TransLPRNet }
  \label{fig24}
\end{figure}

This paper also investigates integrating a Perspective Transformation Network (PTN) with TransLPRNet, leveraging TransLPRNet to provide feedback to the PTN. The corresponding experimental results are shown in Figure \ref{fig24}. The results indicate that the integration of PTN and TransLPRNet can potentially lead to suboptimal rectification performance, and may even exacerbate the geometric distortion of license plate images. This is primarily due to the fact that TransLPRNet is capable of processing license plates with a certain degree of tilt; its performance degrades only under conditions of extreme tilt. Consequently, when TransLPRNet can still effectively recognize the license plate, the PTN struggles to obtain effective feedback adjustment signals, and thus fails to spatially correct the license plate image accurately.
\subsection{TransLPRNet Training and Evaluation}\label{}
To ensure fairness in the subsequent comparative experiments, the TransLPRNet model in this study was trained using the recommended data partitioning strategy from the CCPD dataset \cite{xu2018towards}. The newly constructed multi-type license plate dataset was randomly split into two equal parts—one used for training and validation (with an 8:2 ratio), and the other used for testing. Unlike the original base test set in CCPD, this test set also includes double-line license plates. Therefore, in this study, we further divided it into two subsets: base-s (single-line license plate test set) and base-d (double-line license plate test set).Although the number of samples in the extended single-line and double-line license plate dataset matches that of the original CCPD base training and test sets, the number of single-line plates in both the training and test sets is lower due to redundancy compression and the addition of double-line plates. Despite the reduced quantity, the dataset retains maximal informational value through redundancy compression. As a result, the recognition performance of the trained license plate model remains comparable to that trained on the original CCPD base set. Detailed results are presented in the comparative experiments in Section 4.3.1, and the hyperparameter settings used for training are listed in Table \ref{tbl4}.
\begin{table}[width=.9\linewidth,cols=4,pos=h]
\caption{Hyperparameter configuration for TransLPRNet training}\label{tbl4}
\begin{tabular*}{\tblwidth}{@{} CC@{} }
\toprule
Parameter Name & Number of Parameters\\
\midrule
Learning Rate & 0.0001 \\
Optimizer & Adam \\
Batch size & 32 \\
Training Epochs & 200\\
Image size & 224×224 \\
\bottomrule
\end{tabular*}
\end{table}

\begin{figure}[htbp]
  \centering
  \includegraphics[width=0.6\linewidth]{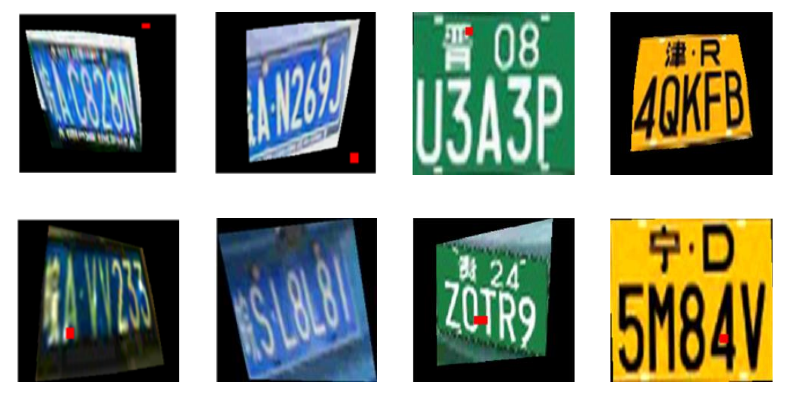}
  \caption{Data augmentation visualization }
  \label{fig25}
\end{figure}

To enhance the model's adaptability in open-world scenarios, this paper incorporates various data augmentation methods into the TransLPRNet model training, thereby expanding the diversity of the training dataset and improving the model's robustness against different scenes and interference factors. The augmentation strategy includes the following techniques: random cropping to simulate variations in license plate position and size; random rotation of license plates to accommodate different viewing angles; color jittering to emulate changes in lighting conditions; random perspective transformation of license plates to mimic camera viewpoint deviations; and random erasing to simulate license plate occlusion or noise. The combination of these data augmentation techniques effectively enriches the training samples, ultimately improving the model's license plate recognition performance in real-world applications. The specific effects are illustrated in Figure \ref{fig25}.
\subsubsection{Comparison of recognition results on the mixed dataset}\label{}
To verify the reliability of TransLPRNet in recognizing both single-line and double-line license plates simultaneously, this paper trains and evaluates TransLPRNet on the mixed dataset constructed in Section 3.3.2. This mixed dataset is built by performing redundancy compression on the Base subset of CCPD and overlaying synthesized double-line license plate images, aiming to evaluate the model's recognition performance when both single-line and double-line license plates coexist. To ensure the fairness of the comparative experiments, this paper selects currently open-source license plate recognition algorithms. The algorithm classification approach and localization error simulation method remain consistent with the previous experimental settings and will not be repeated here. Tables \ref{tb98} and Tables \ref{tb99} show the recognition accuracy, average recognition accuracy, inference speed, and parameter count of each algorithm on the mixed test set under coarse localization perturbation and fine localization perturbation conditions, respectively.

\begin{table}[htbp]
\centering
\small
\setlength{\tabcolsep}{0.5\tabcolsep}
\caption{Performance comparison on the mixed dataset under coarse localization conditions}\label{tb98}
\begin{tabular*}{\linewidth}{@{\extracolsep{\fill}} *{12}{c} @{}}
\toprule
Recognition rate/\% & Avg & Base-s & Base-d & Db & Fn & Rotate & Tilt & Weather & Challenge & FPS & Params \\
\midrule
LPRNet \cite{zherzdev1806lprnet} & 93.15 & 95.85 & × & 97.10 & 97.28 & 90.65 & 94.43 & 96.28 & 80.48  & \textbf{118} & \textbf{0.65} \\
Eulpr \cite{qin2020efficient} & 94.73 & 99.29 & 93.52 & 97.35 & 94.67 & 95.96 & 97.23 & 96.55 & 83.24 & \underline{63} & \underline{1.01} \\
PaddleOCRv3 \cite{li2022pp} & 97.99 & \underline{99.65} & \underline{99.43} & 98.71 & 99.24 & 98.82 & 98.83 & 98.16 & 91.06 & 58 & 8.6 \\
TrOCR+pre-trained weights\cite{li2023trocr} & \textbf{98.82} & 99.64 & \underline{99.43} & \underline{99.24} & \underline{99.32} & \underline{99.37} & \underline{99.45} & \textbf{99.43} & \textbf{94.68} & 12 & 34.6 \\
TrOCR\cite{li2023trocr} & 92.11 & 97.25& 95.63 & 93.18 & 95.18& 93.09 & 93.09&93.87&75.58 & 12 & 34.6 \\
Lpr-transformer \cite{sosopop2024lpr} & 96.64 & 99.30 & 99.34 & 97.86& 98.40& 97.17 & 98.08 & 96.93 & 86.02 & 53 & 1.34 \\
TransLPRNet & \underline{98.75} & \textbf{99.84} & \textbf{99.97} & \textbf{99.28} & \textbf{99.42} & \textbf{99.54} & \textbf{99.70} & \underline{99.12} & \underline{93.14} & 46 & 5.94 \\
\bottomrule
\multicolumn{12}{p{\linewidth}}{Note: “×” indicates that the algorithm does not support double-line license plate recognition. FPS is measured with batch size = 1.}
\end{tabular*}
\end{table}

\begin{table}[htbp]
\centering
\small
\setlength{\tabcolsep}{0.5\tabcolsep}
\caption{Performance comparison on the mixed dataset under fine localization conditions}\label{tb99}
\begin{tabular*}{\linewidth}{@{\extracolsep{\fill}} *{12}{c} @{}}
\toprule
Recognition rate/\% & Avg & Base-s & Base-d & Db & Fn & Rotate & Tilt & Weather & Challenge &  FPS & Params \\
\midrule
LPRNet \cite{zherzdev1806lprnet} & 96.81 & 99.44 & × & 98.24 & 98.58 & 98.71 & 98.74 & 96.98 & 86.97 &  \textbf{118} & \textbf{0.65} \\
Eulpr \cite{qin2020efficient} & 97.23 & 99.48 & 96.32 & 98.75 & 98.69 & 99.07 & 99.09 & 97.74 & 88.68 & \underline{63} & \underline{1.01} \\
PaddleOCRv3 \cite{li2022pp} & 98.20 & 99.65 & 98.43 & 98.82 & 98.91 & 98.78 & 99.01 & 99.07 & 92.93 &  58 & 8.6 \\
TrOCR+pre-trained weights \cite{li2023trocr} & \underline{98.91} & \underline{99.69} & \underline{99.58} & \textbf{99.39} & \textbf{99.51} & \underline{99.75} & \underline{99.67} & \underline{99.32} & \textbf{94.37} & 12 & 34.6 \\
TrOCR\cite{li2023trocr} & 94.95 & 98.51& 98.23 & 95.32& 97.25&94.67& 94.94&94.39& 86.32 & 12 & 34.6 \\
Lpr-transformer \cite{sosopop2024lpr} & 97.62 & 99.53 & 99.51 & 98.21 & 98.76 & 99.57 & 99.37 & 97.34 & 88.64 &  53 & 1.34 \\
TransLPRNet & \textbf{99.03} & \textbf{99.89} & \textbf{99.99} & \underline{99.37} & \underline{99.49} & \textbf{99.91} & \textbf{99.87} & \textbf{99.36} & \underline{94.35} & 46 & 5.94 \\
\bottomrule
\multicolumn{12}{p{\linewidth}}{Note: “×” indicates that the algorithm does not support double-line license plate recognition. FPS is measured with batch size = 1. The fine localization here uses direct correction with labels, which is the same as the algorithms compared above. The reason why TrOCR's accuracy on the Challenge and Weather datasets under fine localization conditions in Table \ref{tb99} is lower than that in Table \ref{tb98} is that TrOCR was trained with occlusion-like data during the pre-training process, resulting in stronger generalization and inference capabilities. However, after correction, since the pre-training process did not involve similar data processing, its performance degrades instead.}
\end{tabular*}
\end{table}

\begin{figure}[htbp]
  \centering
  \includegraphics[width=0.6\linewidth]{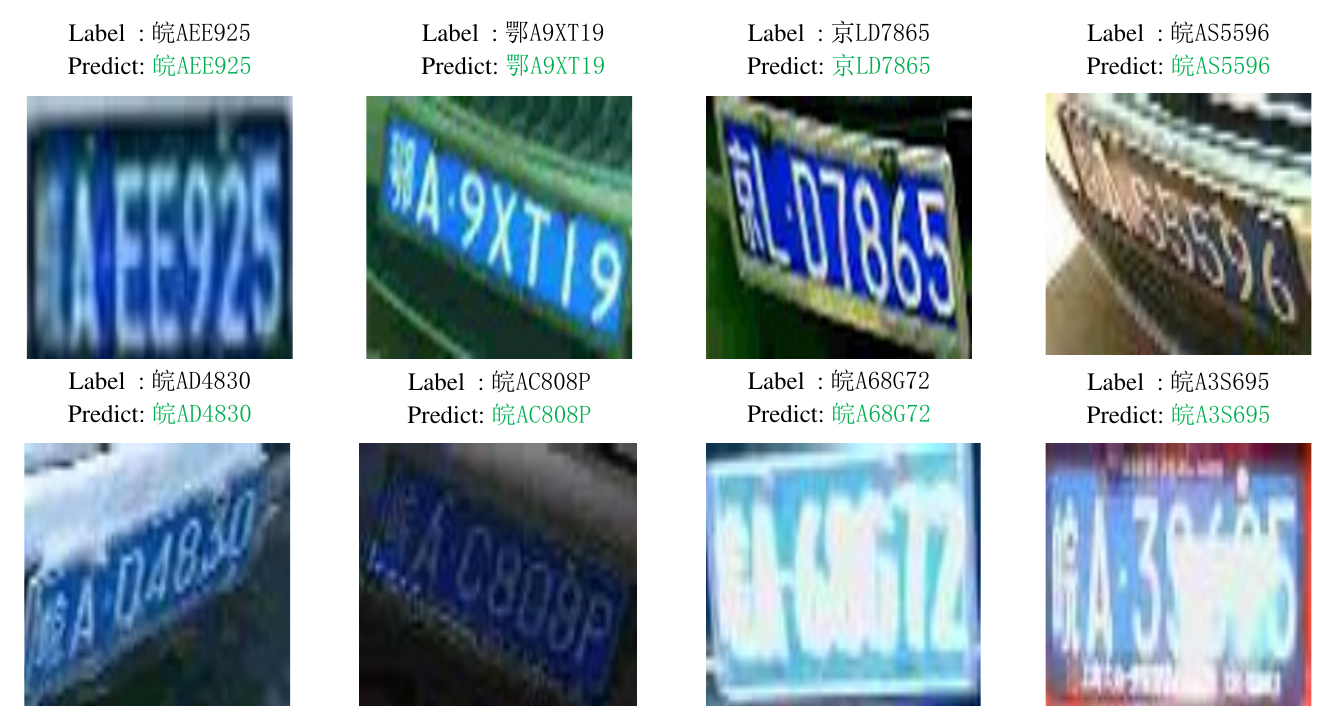}
  \caption{Visualization of single-line license plate image recognition results }
  \label{fig26}
\end{figure}

\begin{figure}[htbp]
  \centering
  \includegraphics[width=0.6\linewidth]{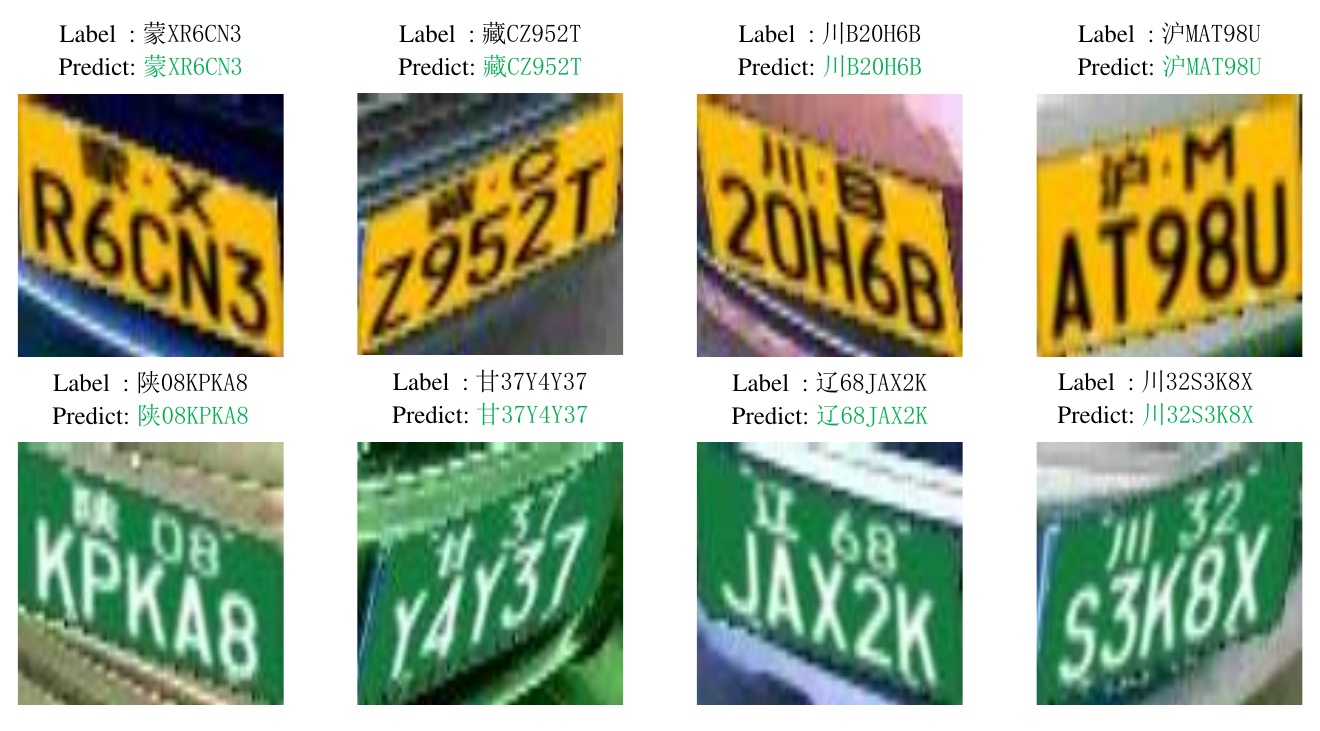}
  \caption{Visualization of double-line license plate image recognition results }
  \label{fig27}
\end{figure}

From Tables \ref{tb98} and Tables \ref{tb99}, it can be seen that Eulpr, PaddleOCR, TrOCR, and the proposed TransLPRNet can all effectively support the task of double-line license plate recognition.From the experimental results, the proposed algorithm outperforms the comparison methods on most subsets under both coarse localization and fine localization conditions, and is only inferior to TrOCR with pre-trained weights on the Challenge and Weather subsets. It should be noted that TrOCR relies on large-scale scene text images for pre-training during its training process, whereas TransLPRNet in this paper is trained entirely from scratch without using any external pre-trained data. Under the same conditions without pre-training, the performance of TrOCR will significantly degrade and cannot achieve the recognition accuracy of the proposed algorithm. This demonstrates that the proposed model achieves excellent recognition performance without relying on pre-trained weights, fully validating the effectiveness and superiority of the TransLPRNet architecture.Figure \ref{fig26} shows the visualization of single-line license plate recognition, and Figure  \ref{fig27} shows the visualization of double-line license plate recognition.

Furthermore, thanks to the combination of the visual encoder and text decoder, TransLPRNet exhibits smaller fluctuations in recognition accuracy under different localization perturbation conditions compared to algorithms such as LPRNet, Eulpr, and PaddleOCRv3, demonstrating stronger stability. Under the condition of batch size = 1, the proposed method achieves an inference speed of 46 FPS, which is much higher than that of TrOCR, while also significantly reducing the number of parameters. These results indicate that the proposed method maintains high recognition accuracy while offering significant advantages in inference speed and parameter count.

\subsection{Ablation experiments }\label{}

To validate the practical effectiveness of the proposed license plate rectification algorithm PTN in enhancing model recognition performance, this study designed three comparative experiments. Specifically, the first experiment involved recognition after applying perturbations to coarsely localized license plates. The second experiment involved recognition after applying PTN rectification to the coarsely perturbed license plates. The third experiment involved recognition following perturbations to finely localized license plates. To ensure the fairness of the comparisons, all experiments employed the same training strategy and training data, with different localization and rectification methods introduced only during the testing phase. The test set includes corrected single-line and double-line license plate samples with previously mislabeled annotations. The detailed comparison results are shown in Table \ref{tbl8}.

\begin{table}[width=.9\linewidth,cols=4,pos=h]
\caption{Comparison of recognition accuracy with different localization methods and before/after PTN correction on the mixed dataset}\label{tbl8}
\begin{tabular*}{\tblwidth}{@{}CCCCCCCCC@{}}
\toprule
Experiment & Avg & Base & Db & Fn & Rotate & Tilt & Weather & Challenge \\
\midrule
Bounding box localization disturbance & 98.60& 99.87 & 99.29 & 99.43 & 99.55 & 99.72 & 99.12 & 93.25 \\
Bounding box localization disturbance+PTN & 98.91 & 99.90 & 99.37 & 99.48 & 99.87 & 99.86 & 99.36 & 94.54 \\
Four Vertex location disturbance & 98.93& 99.92 & 99.38 & 99.50 & 99.92 & 99.87 & 99.37 & 94.57  \\
\bottomrule
\end{tabular*}

\medskip
\noindent\parbox{\tblwidth}{\small Note: Since the license plate results after fine localization perturbation are already relatively upright and show almost no change after PTN correction, no comparison is made here.}
\end{table}

\begin{figure}[htbp]
  \centering
  \includegraphics[width=0.8\linewidth]{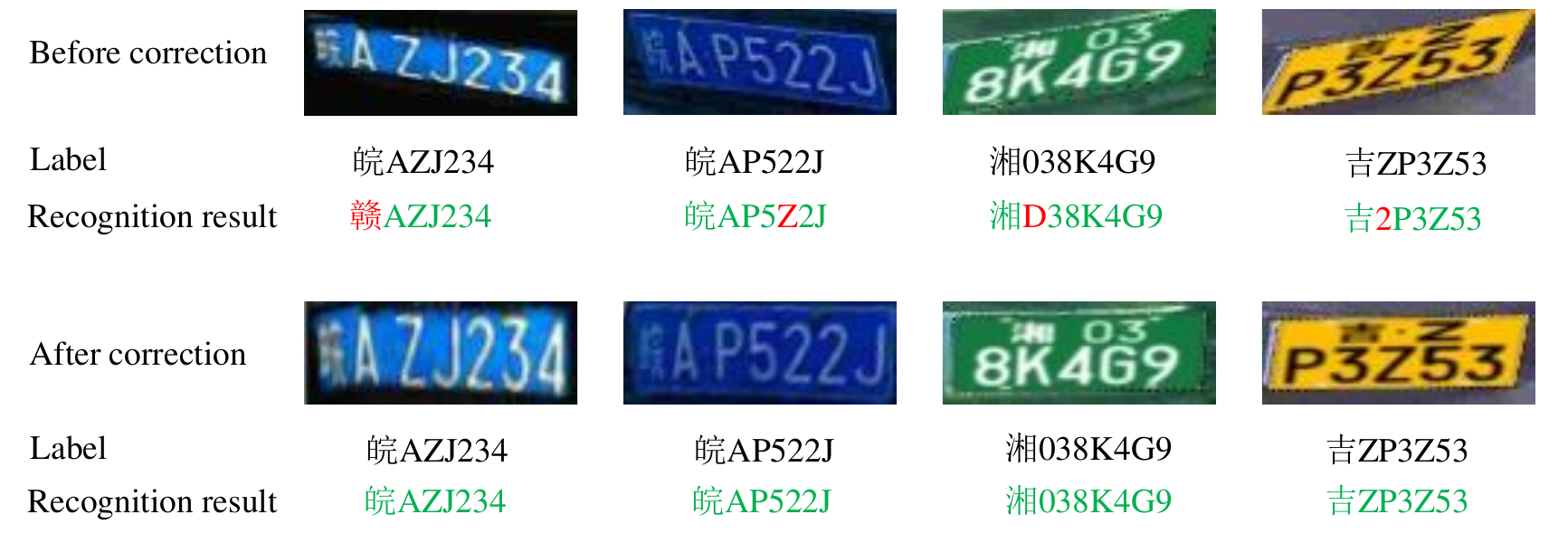}
  \caption{Comparison of recognition results with and without using PTN }
  \label{fig28}
\end{figure}

From the data in Table \ref{tbl8}, it can be seen that the proposed PTN-based correction algorithm effectively improves recognition accuracy under different license plate localization methods and diverse scenarios. In both single-line and double-line license plate test sets, the introduction of the PTN correction module significantly enhances recognition performance across various subsets. After applying PTN correction on single-line license plates with coarse localization perturbation, the recognition accuracy on the Challenge subset increases from 93.25\% to 94.54\%. After applying PTN correction on double-line license plates with coarse localization perturbation, the accuracy improves from 98.60\% to 98.91\%. PTN also demonstrates excellent performance on other subsets, indicating that this module can effectively alleviate recognition difficulties caused by excessive license plate tilt angles. To further validate the effectiveness of PTN, several scene examples were selected for visual analysis, and the results are shown in Figure \ref{fig28}. License plate images without PTN correction exhibit significant tilt and distortion, leading to recognition errors. After PTN correction, the character arrangement of the license plate images becomes more regular and clearer, thereby significantly improving the accuracy and robustness of the recognition model.

\begin{figure}[htbp]
  \centering
  \includegraphics[width=0.6\linewidth]{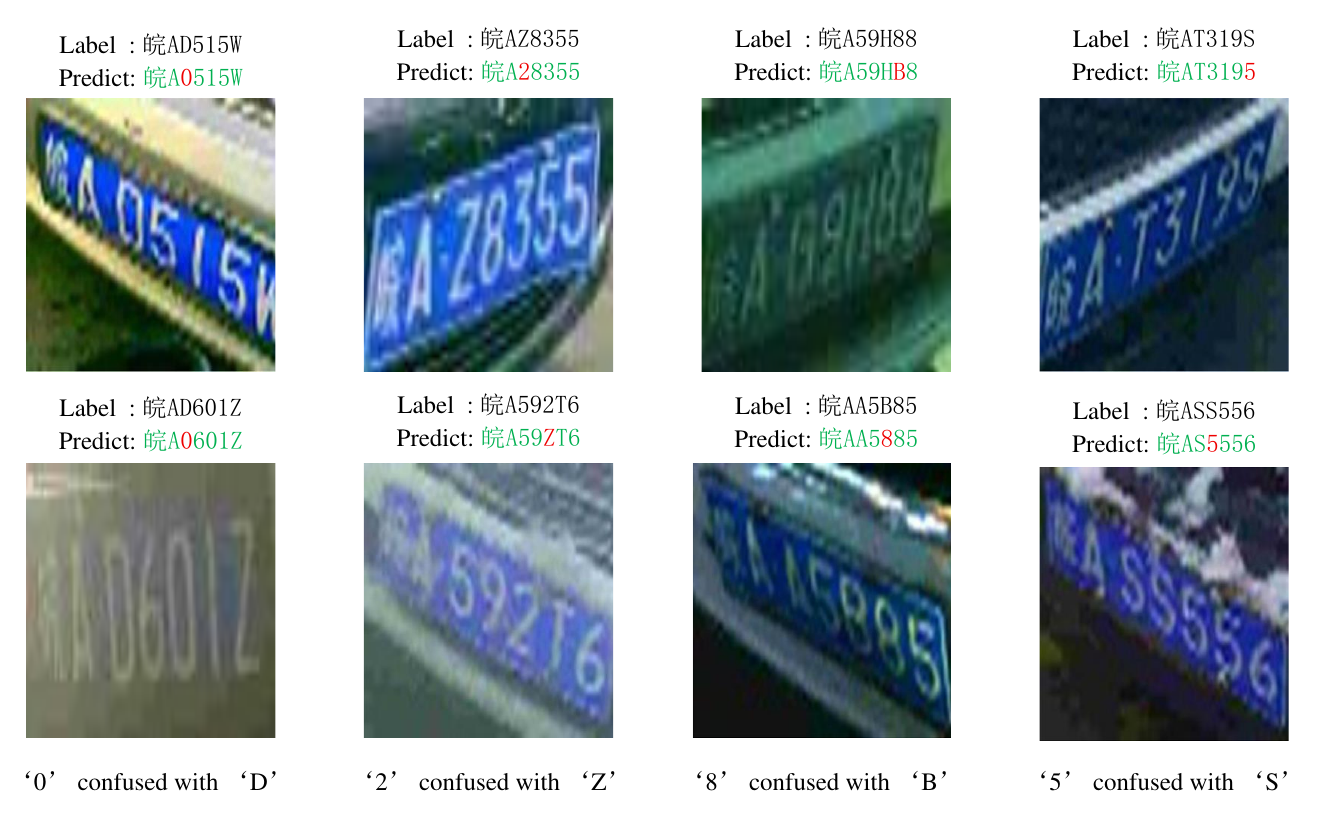}
  \caption{Illustration of license plate character confusion and recognition errors }
  \label{fig30}
\end{figure}

This study analyzes the corrected CCPD test set and finds that the misrecognized license plate images can be mainly categorized into two types. The first type of error stems from the high morphological similarity between characters, manifesting as confusion between the digit "8" and the letter "B", the digit "2" and the letter "Z", the digit "0" and the letter "D", and the digit "5" and the letter "S", as shown in Figure \ref{fig30}. Such recognition errors are caused by the inherent structural similarity of these characters; for example, "8" and "B", as well as "D" and "0", exhibit significant overlap in contour features. In addition, due to various limitations in capturing license plate images in open scenarios — such as low resolution, non-uniform illumination, and character distortion — the key discriminative features of license plate characters are lost, thereby leading to recognition errors.

\begin{figure}[htbp]
  \centering
  \includegraphics[width=0.6\linewidth]{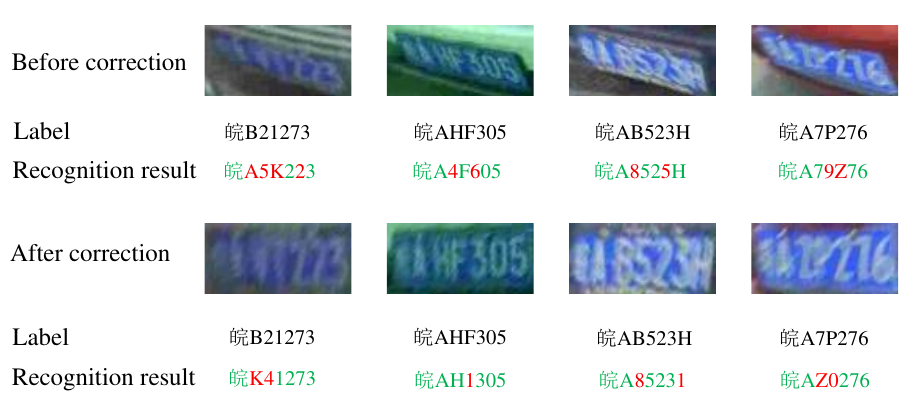}
  \caption{Illustration of license plate character confusion and recognition errors }
  \label{fig31}
\end{figure}

Another factor contributing to license plate recognition errors is the high degree of blurriness inherent to certain license plate images. Such blurriness exceeds the capacity of the human visual system to accurately discern individual characters, resulting in the critical features of each character being substantially compromised. Image quality degradation leads to the loss of fine details in the characters, and even after effective correction using PTN, misrecognition during subsequent processing stages remains a significant challenge. Additionally, for these highly blurred license plates, the Chinese character segment is frequently predicted as “wan,” primarily due to the fact that the CCPD dataset predominantly contains license plates from Anhui Province. This bias during training causes the model to overly associate such characters with “wan.” While this study has attempted to address potential imbalance by augmenting the dataset to improve the distribution of double-line license plates, it has not sufficiently balanced the provincial distribution, which adversely affects the model’s generalization capability, with representative examples provided in Figure \ref{fig31}.
\section{Conclusion and Outlook}\label{}
This paper introduces TransLPRNet, a novel license plate recognition network that integrates a lightweight, pre-trained visual encoder and a text decoder. By employing a Transformer architecture to globally model the inter-relationships between license plate image patches, TransLPRNet effectively mitigates the character loss or spurious character addition issues commonly encountered in CNN+CTC and CNN+RNN based license plate recognition algorithms. Furthermore, in conjunction with the proposed Perspective Transformation Network (PTN), TransLPRNet achieves accurate and rapid recognition of single-line and double-line Chinese license plates in a variety of complex scenarios.

Beyond providing training supervision for the PTN, the proposed front-view license plate classification network also exhibits potential for license plate image quality assessment and fine-grained license plate type classification. This information can be leveraged to enhance the confidence estimation of license plate recognition results, aiding in the determination of result correctness. In future work, we plan to explore the adoption of a Transformer encoder-only architecture in place of the current Transformer encoder-decoder structure within TransLPRNet, with the potential to further improve inference speed.

\clearpage %%Remove this from your manuscript

% Uncomment and use as the case may be
%\begin{theorem} 
%\end{theorem}

% Uncomment and use as the case may be
%\begin{lemma} 
%\end{lemma}


\printcredits
\bibliographystyle{cas-model2-names}

\bibliography{cas-refs}

@article{shi2023license,
  title={License plate recognition system based on improved YOLOv5 and GRU},
  author={Shi, Hengliang and Zhao, Dongnan},
  journal={Ieee Access},
  volume={11},
  pages={10429--10439},
  year={2023},
  publisher={IEEE}
}

@article{fan2022improving,
  title={Improving robustness of license plates automatic recognition in natural scenes},
  author={Fan, Xudong and Zhao, Wei},
  journal={IEEE Transactions on Intelligent Transportation Systems},
  volume={23},
  number={10},
  pages={18845--18854},
  year={2022},
  publisher={IEEE}
}

@article{liu2019vehicle,
  title={Vehicle license plate recognition method based on deep convolution network in complex road scene},
  author={Liu, Ze and Cai, Yingfeng and Chen, Long and Wang, Hai and He, Youguo},
  journal={Proceedings of the Institution of Mechanical Engineers, Part D: Journal of Automobile Engineering},
  volume={233},
  number={9},
  pages={2284--2292},
  year={2019},
  publisher={SAGE Publications Sage UK: London, England}
}

@article{huang2020new,
  title={A new approach for character recognition of multi-style vehicle license plates},
  author={Huang, Qiuying and Cai, Zhanchuan and Lan, Ting},
  journal={IEEE Transactions on multimedia},
  volume={23},
  pages={3768--3777},
  year={2020},
  publisher={IEEE}
}

@article{he2020robust,
  title={Robust automatic recognition of Chinese license plates in natural scenes},
  author={He, Ming-Xiang and Hao, Peng},
  journal={Ieee Access},
  volume={8},
  pages={173804--173814},
  year={2020},
  publisher={IEEE}
}

@article{zou2022license,
  title={License plate detection and recognition based on YOLOv3 and ILPRNET},
  author={Zou, Yongjie and Zhang, Yongjun and Yan, Jun and Jiang, Xiaoxu and Huang, Tengjie and Fan, Haisheng and Cui, Zhongwei},
  journal={Signal, image and video processing},
  volume={16},
  number={2},
  pages={473--480},
  year={2022},
  publisher={Springer}
}

@article{zou2020robust,
  title={A robust license plate recognition model based on bi-lstm},
  author={Zou, Yongjie and Zhang, Yongjun and Yan, Jun and Jiang, Xiaoxu and Huang, Tengjie and Fan, Haisheng and Cui, Zhongwei},
  journal={IEEE Access},
  volume={8},
  pages={211630--211641},
  year={2020},
  publisher={IEEE}
}

@article{zherzdev1806lprnet,
  title={Lprnet: License plate recognition via deep neural networks. arXiv 2018},
  author={Zherzdev, S and Gruzdev, A},
  journal={arXiv preprint arXiv:1806.10447}
}

@article{hua2024recognition,
  title={Recognition of vehicle license plates in highway scenes with deep fusion network and connectionist temporal classification},
  author={Hua, Liru and Ma, Xinyi and Zhao, Chihang and Zhang, Bailing and Su, Zijun and Wu, Yuhang},
  journal={IET Image Processing},
  volume={18},
  number={13},
  pages={4066--4080},
  year={2024},
  publisher={Wiley Online Library}
}

@inproceedings{raj2022license,
  title={License plate recognition system using yolov5 and cnn},
  author={Raj, Shreya and Gupta, Yash and Malhotra, Ruchika},
  booktitle={2022 8th International Conference on Advanced Computing and Communication Systems (ICACCS)},
  volume={1},
  pages={372--377},
  year={2022},
  organization={IEEE}
}

@article{adak2022automatic,
  title={Automatic number plate recognition (ANPR) with YOLOv3-CNN},
  author={Adak, Rajdeep and Kumbhar, Abhishek and Pathare, Rajas and Gowda, Sagar},
  journal={arXiv preprint arXiv:2211.05229},
  year={2022}
}

@article{jaderberg2015spatial,
  title={Spatial transformer networks},
  author={Jaderberg, Max and Simonyan, Karen and Zisserman, Andrew and others},
  journal={Advances in neural information processing systems},
  volume={28},
  year={2015}
}

@article{deng2024collaborative,
  title={Collaborative License Plate Recognition via Association Enhancement Network With Auxiliary Learning and a Unified Benchmark},
  author={Deng, Yifei and Wang, Guohao and Li, Chenglong and Wang, Wei and Zhang, Cheng and Tang, Jin},
  journal={IEEE Transactions on Multimedia},
  year={2024},
  publisher={IEEE}
}

@article{yang2024deep,
  title={A Deep Learning-based Framework for Vehicle License Plate Detection.},
  author={Yang, Deming and Yang, Ling},
  journal={International Journal of Advanced Computer Science \& Applications},
  volume={15},
  number={1},
  year={2024}
}

@inproceedings{li2023trocr,
  title={Trocr: Transformer-based optical character recognition with pre-trained models},
  author={Li, Minghao and Lv, Tengchao and Chen, Jingye and Cui, Lei and Lu, Yijuan and Florencio, Dinei and Zhang, Cha and Li, Zhoujun and Wei, Furu},
  booktitle={Proceedings of the AAAI conference on artificial intelligence},
  volume={37},
  number={11},
  pages={13094--13102},
  year={2023}
}

@article{wadekar2022mobilevitv3,
  title={Mobilevitv3: Mobile-friendly vision transformer with simple and effective fusion of local, global and input features},
  author={Wadekar, Shakti N and Chaurasia, Abhishek},
  journal={arXiv preprint arXiv:2209.15159},
  year={2022}
}

@inproceedings{xu2018towards,
  title={Towards end-to-end license plate detection and recognition: A large dataset and baseline},
  author={Xu, Zhenbo and Yang, Wei and Meng, Ajin and Lu, Nanxue and Huang, Huan and Ying, Changchun and Huang, Liusheng},
  booktitle={Proceedings of the European conference on computer vision (ECCV)},
  pages={255--271},
  year={2018}
}

@article{qin2020efficient,
  title={Efficient and unified license plate recognition via lightweight deep neural network},
  author={Qin, Shuxin and Liu, Sijiang},
  journal={IET Image Processing},
  volume={14},
  number={16},
  pages={4102--4109},
  year={2020},
  publisher={Wiley Online Library}
}

@article{du2020pp,
  title={Pp-ocr: A practical ultra lightweight ocr system},
  author={Du, Yuning and Li, Chenxia and Guo, Ruoyu and Yin, Xiaoting and Liu, Weiwei and Zhou, Jun and Bai, Yifan and Yu, Zilin and Yang, Yehua and Dang, Qingqing and others},
  journal={arXiv preprint arXiv:2009.09941},
  year={2020}
}

@article{li2022pp,
  title={PP-OCRv3: More attempts for the improvement of ultra lightweight OCR system},
  author={Li, Chenxia and Liu, Weiwei and Guo, Ruoyu and Yin, Xiaoting and Jiang, Kaitao and Du, Yongkun and Du, Yuning and Zhu, Lingfeng and Lai, Baohua and Hu, Xiaoguang and others},
  journal={arXiv preprint arXiv:2206.03001},
  year={2022}
}

@article{mehta2021mobilevit,
  title={Mobilevit: light-weight, general-purpose, and mobile-friendly vision transformer},
  author={Mehta, Sachin and Rastegari, Mohammad},
  journal={arXiv preprint arXiv:2110.02178},
  year={2021}
}

@article{khokhar2024integrating,
  title={Integrating YOLOv8 and CSPBottleneck based CNN for enhanced license plate character recognition},
  author={Khokhar, Sahil and Kedia, Deepak},
  journal={Journal of Real-Time Image Processing},
  volume={21},
  number={5},
  pages={168},
  year={2024},
  publisher={Springer}
}

@article{xiao2021robust,
  title={Robust license plate detection and recognition with automatic rectification},
  author={Xiao, Degui and Zhang, Lu and Li, Jianfang and Li, Jiazhi},
  journal={Journal of Electronic Imaging},
  volume={30},
  number={1},
  pages={013002--013002},
  year={2021},
  publisher={Society of Photo-Optical Instrumentation Engineers}
}

@inproceedings{bakshi2023alpr,
  title={ALPR-An Intelligent Approach Towards Detection and Recognition of License Plates in Uncontrolled Environments},
  author={Bakshi, Akshay and Gulhane, Sudhanshu and Sawant, Tanish and Sambhe, Vijay and Udmale, Sandeep S},
  booktitle={International Conference on Distributed Computing and Intelligent Technology},
  pages={253--269},
  year={2023},
  organization={Springer}
}

@article{zhang2020robust,
  title={A robust attentional framework for license plate recognition in the wild},
  author={Zhang, Linjiang and Wang, Peng and Li, Hui and Li, Zhen and Shen, Chunhua and Zhang, Yanning},
  journal={IEEE Transactions on Intelligent Transportation Systems},
  volume={22},
  number={11},
  pages={6967--6976},
  year={2020},
  publisher={IEEE}
}

@article{wang2022lsv,
  title={LSV-LP: Large-scale video-based license plate detection and recognition},
  author={Wang, Qi and Lu, Xiaocheng and Zhang, Cong and Yuan, Yuan and Li, Xuelong},
  journal={IEEE Transactions on Pattern Analysis and Machine Intelligence},
  volume={45},
  number={1},
  pages={752--767},
  year={2022},
  publisher={IEEE}
}

@inproceedings{howard2019searching,
  title={Searching for mobilenetv3},
  author={Howard, Andrew and Sandler, Mark and Chu, Grace and Chen, Liang-Chieh and Chen, Bo and Tan, Mingxing and Wang, Weijun and Zhu, Yukun and Pang, Ruoming and Vasudevan, Vijay and others},
  booktitle={Proceedings of the IEEE/CVF international conference on computer vision},
  pages={1314--1324},
  year={2019}
}

@inproceedings{zhang1999flexible,
  title={Flexible camera calibration by viewing a plane from unknown orientations},
  author={Zhang, Zhengyou},
  booktitle={Proceedings of the seventh ieee international conference on computer vision},
  volume={1},
  pages={666--673},
  year={1999},
  organization={Ieee}
}

@misc{sosopop2024lpr,
  author       = {sosopop},
  title        = {chinese-lpr-transformer},
  year         = {2024},
  howpublished = {GitHub},
  url          = {https://github.com/sosopop/chinese-lpr-transformer},

}



\end{document}